\documentclass[letterpaper]{article} 
\usepackage{aaai25}
\usepackage{times}  
\usepackage{helvet}  
\usepackage{courier}  
\usepackage[hyphens]{url}  
\usepackage{graphicx} 
\usepackage{natbib}  
\usepackage{caption} 
\usepackage{algorithm}
\usepackage[table]{xcolor}

\usepackage{xr}
\makeatletter
\newcommand*{\addFileDependency}[1]{
  \typeout{(#1)}
  \@addtofilelist{#1}
  \IfFileExists{#1}{}{\typeout{No file #1.}}
}
\makeatother

\usepackage{newfloat}
\usepackage{listings}
\DeclareCaptionStyle{ruled}{labelfont=normalfont,labelsep=colon,strut=off} 
\floatstyle{ruled}
\newfloat{listing}{tb}{lst}{}
\floatname{listing}{Listing}
\definecolor{yescolor}{RGB}{0, 0, 0}     
\definecolor{partialcolor}{RGB}{0, 0, 0} 
\definecolor{nocolor}{RGB}{0, 0, 0}        
\definecolor{nacolor}{RGB}{0, 0, 0}    

\usepackage{amsfonts}       
\usepackage{nicefrac}  
\usepackage{booktabs} 
\usepackage{multirow}
\usepackage{booktabs} 
\usepackage{adjustbox}
\usepackage{subcaption}
\usepackage{float}
\usepackage{longtable}

\usepackage{microtype}
\usepackage{graphicx}

\usepackage{caption}
\usepackage{amssymb}
\usepackage{amsfonts}
\usepackage{amsmath}
\usepackage{amsthm}
\usepackage{xr}

\usepackage{algorithm}
\usepackage{algpseudocode}
\usepackage{mdframed}

\newtheorem{theorem}{Theorem}
\newtheorem{lemma}{Lemma}
\newtheorem{proposition}[theorem]{Proposition}
\newtheorem{remark}{Remark}
\newtheorem{corollary}[theorem]{Corollary}
\newtheorem{claim}{Claim}
\newtheorem{definition}{Definition}
\newtheorem{assumption}{Assumption}
\usepackage{threeparttable}

\title{Federated Unsupervised Domain Generalization using Global and Local Alignment of Gradients}
\author {
    Farhad Pourpanah\equalcontrib, 
    Mahdiyar Molahasani\equalcontrib, 
    Milad Soltany\equalcontrib,  
    Michael Greenspan, 
    Ali Etemad 
}
\affiliations {
    Queen’s University, Canada\\
    \{f.pourpanahnavan, m.molahasani, milad.soltany, michael.greenspan, ali.etemad\}@queensu.ca
}

\usepackage{bibentry}

\begin{document}

\maketitle

\begin{abstract}
We address the problem of federated domain generalization in an unsupervised setting for the first time. We first theoretically establish a connection between domain shift and alignment of gradients in unsupervised federated learning and show that aligning the gradients at both client and server levels can facilitate the generalization of the model to new (target) domains. Building on this insight, we propose a novel method named FedGaLA, which performs gradient alignment at the client level to encourage clients to learn domain-invariant features, as well as global gradient alignment at the server to obtain a more generalized aggregated model. To empirically evaluate our method, we perform various experiments on four commonly used multi-domain datasets, PACS, OfficeHome, DomainNet, and TerraInc. The results demonstrate the effectiveness of our method which outperforms comparable baselines. Ablation and sensitivity studies demonstrate the impact of different components and parameters in our approach. The source code is available at: https://github.com/MahdiyarMM/FedGaLA.
\end{abstract}

%

\section{Introduction}
\label{sec:intro}
Federated learning \cite{mcmahan2017communication,zhang2021survey} has emerged as a promising framework for training machine learning models across multiple decentralized clients while preserving data privacy. It allows clients to collaboratively train a global model without the need to exchange their sensitive and local data. Each client trains a local model using its data and a server aggregates these models at a certain frequency \cite{ghosh2020efficient,charles2021large}. However, given that each client collects a different set of local training data, two issues arise. First, the data collected by each client is often recorded under unique conditions that may result in mutual domain shifts \cite{liu2021feddg}. Second, labeling training data is inherently challenging and resource-intensive; this issue is even more pronounced in the context of federated settings. 
A typical example of this scenario is a network of wearable activity monitors where variations in user conditions such as demographics or ambient factors can lead to significant domain shifts across devices, meanwhile, the users are generally not asked to provide ground-truth labels for their performed activities.

In prior works, each of these issues has been addressed as a separate problem statement: (\textit{i}) \emph{federated domain generalization} \cite{zhang2021federated,nguyen2022fedsr,bai2024benchmarking}, and (\textit{ii}) \emph{federated unsupervised learning} \cite{zhuang2021collaborative,han2022fedx}. Despite the effectiveness of both problem definitions, each ignores the fundamental assumptions of the other regarding the data in terms of distributions and availability of labels.
To further approach federated learning in a more practical scenario, we propose to merge these two under a new umbrella called \emph{federated unsupervised domain generalization}, which we define as Definition \ref{def1}. To our knowledge, prior works have not studied federated learning under such constraints.

\begin{definition}\label{def1}
Federated unsupervised domain generalization is the problem of learning general representations from various decentralized \textbf{unlabeled} datasets, each belonging to a \textbf{different domain}, in a federated setup where data sharing is restricted due to privacy concerns.
\end{definition}

\begin{figure}[t]
    \centering
    \includegraphics[width=\linewidth]{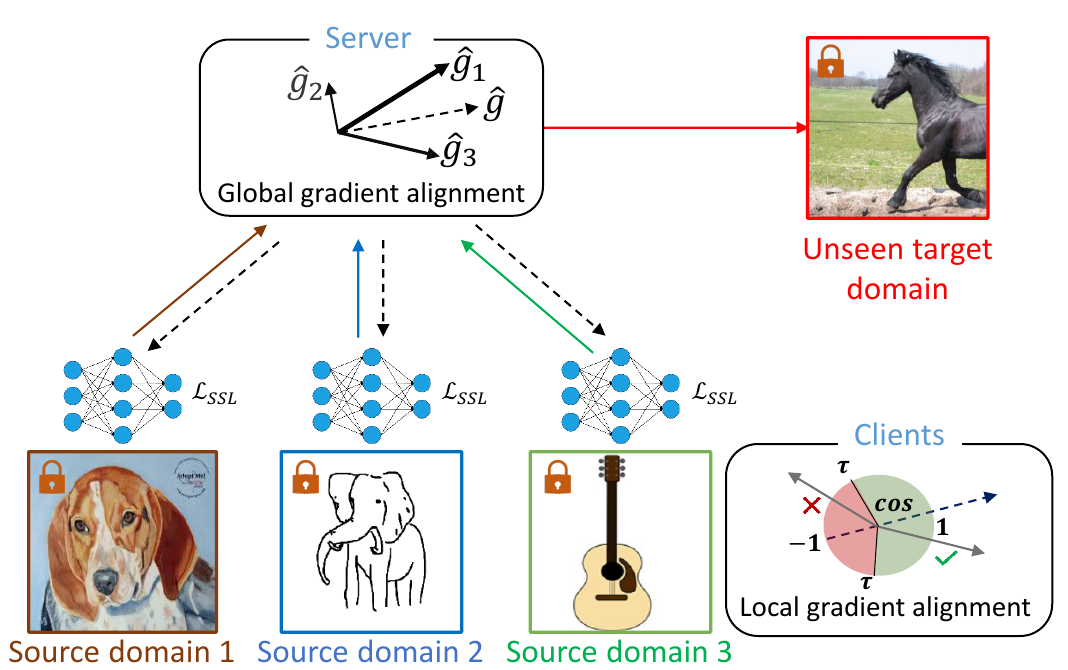}
    \caption{Overview of \emph{FedGaLA}.}
    \label{fig:overview}
    \end{figure}
To address this new problem, we propose a novel method called \textit{\textbf{Fed}erated Unsupervised Domain Generalization using \textbf{G}lobal \textbf{a}nd \textbf{L}ocal \textbf{A}lignment of Gradients (FedGaLA)}. 
To learn more generalized representations from multiple domains, FedGaLA relies on gradient alignment at both client (local) and server (global) levels. At the client level, local models are trained with unlabeled local data available to each client using self-supervised learning (SSL). To learn domain-invariant representations, gradients that are not aligned with the reference gradient, i.e., the global learning direction, are discarded. At the server level, to achieve better generalization, the local models are aggregated based on their alignment with each other (see Figure \ref{fig:overview}). Specifically, the local gradients that are more closely aligned with the average gradient are given greater weight in the aggregation stage. 
Moreover, we provide a detailed theoretical framework establishing a connection between the alignment of gradients from different clients and the similarity between their data distributions.
Since FedGaLA does not require the sharing of data between clients to perform alignment at either the server or client sides, it effectively preserves data privacy. We verify the performance of our approach using four public datasets, PACS \cite{yu2022pacs}, OfficeHome \cite{venkateswara2017deep}, DomainNet \cite{peng2019moment}, and TerraInc \cite{beery2018recognition} and demonstrate strong performance in federated unsupervised domain generalization in comparison to various baselines. We also conduct different ablation and sensitivity studies to understand the impact of different parameters and the choice of SSL frameworks.

In summary, the main contributions of this study are as follows.
(\textbf{1}) To our knowledge, this is the first work to address federated unsupervised domain generalization, introducing a new problem class in this area.
(\textbf{2}) We propose a novel technique that aligns gradients at both the local and global levels. In doing so, our solution effectively extracts domain-invariant information in local training and aligns each client's contributions during the aggregation process, boosting the model's generalization across various target domains. (\textbf{3}) We conduct extensive experiments and ablations to demonstrate the effectiveness of our proposed method on various benchmarks. To enable fast reproducibility and contribute to the area, we make our code public at: \url{https://github.com/MahdiyarMM/FedGaLA}.


\section{Related work}
\label{sec:related}

\noindent \textbf{Federated learning.} 
Federating learning is a technique of distributed training that enables learning from decentralized clients without the need to transfer raw data to a central server \cite{mcmahan2017communication, zhang2021survey}. In the first work in the area of federated learning, FedAVG~\cite{mcmahan2017communication} aggregates the weights of clients via averaging to form the global model. Various federated learning techniques have been since developed for different purposes. FedProx \cite{li2020federated} adds a proximal term to the objective to tackle the data heterogeneity problem. FedBuf \cite{nguyen2022federated} proposes to use buffered asynchronous aggregation to address scalability and privacy issues. FedALA \cite{zhang2023fedala} dynamically aggregates the downloaded global model with the local model on each client, ensuring alignment with the local objective. Moon \cite{li2021model} leverages the similarity between model representations to improve the local training of individual parties through contrastive learning. 
FLTrust \cite{cao2020fltrust} and SignGuard \cite{xu2022byzantine} leverage gradient filtering to detect and eliminate malicious gradients with the aim of enhancing the robustness of federated learning systems against model poisoning attacks.

\noindent \textbf{Federated domain generalization.} This category of problems aims to learn a general model in a privacy-preserving manner that generalizes well to new target domains with distribution shifts \cite{liu2021feddg,li2023federated}. There have been several studies to explore this direction, mainly focusing on learning domain-invariant features \cite{nguyen2022fedsr,wu2021collaborative} or identifying common features across multiple domains \cite{zhang2021federated}. FedDG \cite{liu2021feddg} allows clients to share their data in the frequency space for medical image segmentation. FedSR \cite{nguyen2022fedsr} proposes to learn a simple representation of data to better generalize to target domains through two regularizers, i.e., L2-norm and conditional Mutual Information. FedADG \cite{zhang2021federated} leverages adversarial training to align source domain distributions by matching each distribution with a reference distribution. FedKA \cite{sun2023feature} learns domain-invariant features by employing feature distribution matching in a universal workspace. CCST \cite{chen2023federated} aligns various client distributions and mitigates model biases by adapting local models to diverse sample styles via cross-client style transfer. FedDG-GA \cite{zhang2023federatedGA} uses domain divergence and a moment mechanism to enhance generalization through dynamic domain weight adjustment.

FedSB \cite{soltany2024federated} enhances domain generalization by introducing controlled uncertainty to local clients and ensuring balanced contributions from each client.
StableFDG \cite{park2024stablefdg} uses style and attention-based strategies to address the federated domain generalization problem. hFedF \cite{bartholet2024non} utilizes hypernetworks for non-linear aggregation to facilitate generalization to unseen domains. PerAda \cite{xie2024perada} uses knowledge distillation to align each client’s adapter with a global adapter, blending their knowledge. This approach helps the model handle shifting distributions across diverse domains while keeping computational costs low. gPerXAN \cite{le2024efficiently} combines a personalized normalization scheme with a guiding regularizer to address domain shifts in federated learning, reducing privacy risks and communication costs. MCGDM \cite{wei2024multi} uses intra- and inter-domain gradient matching to minimize domain-specific overfitting, promoting robust generalization on unseen domains. Finally, FGGP \cite{wan2024federated} captures domain information using clustered prototypes and contrastive learning.

\noindent \textbf{Federated unsupervised learning.} 
This category attempts to learn representations from unlabeled data distributed across clients while preserving data privacy \cite{jin2020towards}. The study by \cite{van2020towards} is the first to propose federated unsupervised learning using an encoder-decoder structure. FedU \cite{zhuang2021collaborative} employs a contrastive learning approach with online and target networks to enable each client to learn representations from unlabeled data independently. It also introduces a dynamic aggregation mechanism to update the predictor, either locally or globally. Similarly, FedEMA \cite{zhuang2022divergence} uses the exponential moving average of the global model to update local ones. FedX \cite{han2022fedx} proposes to use knowledge distillation to learn representation from local data and refine the central server’s knowledge. FedCA \cite{zhang2023federated} consists of two components: (\textit{i}) a dictionary module to gather the representations of samples from each client to maintain consistency in the representation space across all clients, and (\textit{ii}) an alignment module to adjust each client's representation to match a base model trained on public data.

\section{Approach}
\label{sec:approach}

\subsection{Problem formulation}
\label{secsec:problem}

To formalize Definition \ref{def1}, assume $K$ clients, $C_i$, in a federated setup, each with its own \textit{unlabeled} data $\mathcal{D}_i = \{ \textbf{x}_i^{(n)} \}_{n=1}^{N_i}$. Each dataset consists of $N_i$ data points 
sampled from a distinct data distribution $p(\textbf{x}_i)$, where $\textbf{x}_i$ is a vector of $F$ features, i.e., $\textbf{x}_i = [x_i^1, x_i^2, ... x_i^F]^T$. The data distributions are assumed to be different among the clients with each distribution $p(\textbf{x}_i)$ sampled from a family of distributions $\mathcal{P}$. Privacy constraints prevent the transfer of data between clients or to the server $S$. The objective is to learn generalized representations from  $\textbf{x}_i$ that perform well across unseen distributions $p(\textbf{x}_t) \sim \mathcal{P}$, where $p(\textbf{x}_i) \neq p(\textbf{x}_t)$. This is formulated as minimizing the expected loss over the unseen distributions: 
\begin{equation}
 \label{Eq.Global}
    \min_{\theta} \mathbb{E}_{p(\textbf{x}_t) \sim \mathcal{P}} \left[ \mathbb{E}_{p(\textbf{x}_t)} \left[ \mathcal{L}(\theta; \textbf{x}) \right] \right],
\end{equation}
where $\mathcal{L}$ is the unsupervised loss function, and $\theta$ is the set of global model parameters. Each client contributes to this goal by computing a local objective function approximating the expected loss with respect to its own data distribution:
\begin{equation}
 \min_{\theta_i} \mathbb{E}_{p(\textbf{x}_i)} \left[ l_i(\theta_i; \textbf{x}) \right] \approx \frac{1}{N_i}\sum^{N_i}_{n=1}\mathcal{L}(\theta_i; \textbf{x}_i^{(n)}),
\end{equation}
where $\theta_i$ indicates the local parameters of client $C_i$, and $\theta$ is the global aggregation of all local $\theta_i$.

\subsection{Gradient alignment and domain shift}
\label{secsec:motiv}
  
Under a federated learning framework, privacy constraints prevent clients and servers from accessing each other's data, including distribution information such as data means and variances. They can, however, observe individual client gradients at the server level and the average aggregated gradient across clients at the client level. 
We motivate our work on the fact that alignment of gradients may infer characteristics of the client domain distributions, thus facilitating improved model generalization. 
While empirically the utility of gradients has been demonstrated in the area of domain generalization \cite{mansilla2021domain}, no theoretical basis has been proposed for this approach under federated constraints.
Accordingly, we aim to establish a link between gradient alignment and domain shifts within the proposed problem formulation (Definition \ref{def1}). Our theoretical findings provide the basis for effective local parameter updates and global model aggregation to address federated unsupervised domain generalization.

\begin{assumption}\label{assumption1}
 Let each \emph{$\textbf{x}_i^f$} be a random variable drawn from a Normal distribution \emph{$p(\textbf{x}_i^f) \sim \mathcal{P}$}~\cite{bar2003learning}. Within a single domain, following \cite{vstrumbelj2014explaining,vstrumbelj2011general},  features are assumed to be independent (\emph{$Cov(x_i^{f_1},x_i^{f_2})=0$}). Across different domains, corresponding features of \emph{$\textbf{x}_i^f$} and \emph{$\textbf{x}_j^f$} are bivariate with a covariance of \emph{$\sigma_{x_i^f,x_j^f}$}.
 In line with contemporary practices in deep learning, the features are normalized with \emph{$\mu=0$} and  \emph{$\sigma^2=1$} \cite{yu2022normalization, qi2022note, huang2023normalization}.
 We also assume that each client is a one-layer encoder with the sigmoid activation function. Following \cite{chen2020simple}, 
local training is performed using contrastive loss where a random augmentation of a data sample is used as the positive pair and all other samples are utilized as the negative pairs. The random augmentation is performed with the same random Affine transformation (\emph{$Ax_i+B$}) \cite{wang2023learning} broadcast to all clients. After each epoch, the local models are aggregated in the server and sent back to all clients. The gradient \emph{$\textbf{g}_i$} of the model is assumed to be differentiable.
\end{assumption}

\begin{theorem}[Gradient Misalignment in  Federated Self-supervised Learning Dependent upon Domain Shift]\label{theorem}
Given Assumption \ref{assumption1}, under the problem proposed in Definition \ref{def1}, for two distinct domains characterized by random variables \emph{$\textbf{x}_i$} and \emph{$\textbf{x}_j$} belonging to two different clients $C_i$ and \emph{$C_j$}, an increase in domain shift across the clients results in a decrease in covariance  
$ \text{Cov}(\emph{$\textbf{g}_i$},\emph{$\textbf{g}_j$})$ of the corresponding gradients 
\emph{$\textbf{g}_i$},
\emph{$\textbf{g}_j$}
across \emph{$C_i$} and \emph{$C_j$}'s respective local models.
\end{theorem}

\begin{proof}[Proof of Theorem 1]
Motivated by previous works where the similarity between the representations of different domains under domain shift is measured by Mutual Information 
\cite{gao2020reducing, menapace2020learning,wang2021learning}, we use this concept for modeling the similarity between different domains drawn from a family of distributions. 
To calculate the mutual information we introduce the following Lemma.

\setcounter{lemma}{0}
\begin{lemma}\label{lemma_0}
Given Assumption \ref{assumption1}, the mutual information between two random variables \emph{$\textbf{x}_i$} and \emph{$\textbf{x}_j$} can be calculated as:
\begin{equation}\label{eq:lemma0}
I(\textbf{\emph{x}}_i; \textbf{\emph{x}}_j) = -\frac{1}{2}\sum_{f=1}^F\log(1-\sigma_{x_i^f,x_j^f}^2).
\end{equation}
The full proof of this lemma is presented in Appendix \ref{proof_lemma_0}.
\end{lemma}
Since $\sigma_{x_i^f,x_i^f}=\sigma_{x_i^f}^2$, given identical and standardized domains we have $\sigma_{x_i^f,x_j^f} = 1$. 
Therefore, we observe from Eq. (\ref{eq:lemma0}) that as the shift between the domain approaches zero, the Mutual Information approaches infinity. On the other hand, as the two domains shift apart, $\sigma_{x_i^f,x_j^f}$ approaches zero, and consequently the Mutual information monotonically decreases toward zero. Accordingly, $I(\textbf{x}_i; \textbf{x}_j)$ and $\sigma_{x_i^f,x_i^f}$  are positively and monotonically correlated. To establish the link between the covariance of the features and the variance of the gradient and demonstrate their relationship, the following lemma and claim are introduced.

\begin{lemma}\label{lemma_1}
Given Assumption \ref{assumption1}, the covariance between the differentiable function $\textbf{\emph{g}}$ with inputs \emph{$\textbf{x}_i$} and \emph{$\textbf{x}_j$} can be estimated as:
\begin{equation}\label{eq:lemma2}
    \emph{\text{Cov}}(\emph{\textbf{g}}_i,\textbf{\emph{g}}_j)_{mn} \approx \sum_{f=1}^F\sigma_{x^f_i,x^ f_j}\left(\frac{\partial g_{i(m)}}{\partial x_i^f}\bigg|_{\mu_i^f}\right)
                            \left(\frac{\partial g_{j(n)}}{\partial x_j^f}\bigg|_{\mu_j^f}\right),
\end{equation}
where \emph{$\sigma_{x_i^f,x_j^f}$} is the covariance between the \emph{$f^{th}$} feature of \emph{$x_i$} and \emph{$x_j$} and ${g}_{i(m)}$ is the $m^{th}$ dimension of $g_{i}$. The proof for this Lemma is provided through the Taylor expansion derived in Appendix \ref{Ap:lem2}. 
\end{lemma}

\begin{claim} \label{claim2}
For all clients trained under the federated unsupervised domain generalization setup using self-supervised learning described in Assumption \ref{assumption1}, for two clients \emph{$C_i$} and $C_j$, for both positive and negative contrastive data pairs, the sign of $(\frac{\partial \textbf{\emph{g}}_i}{\partial x^f_i}\bigg|_{x^f_i=\mu^f_i})(\frac{\partial \textbf{\emph{g}}_j}{\partial x^f_j}\bigg|_{x^f_j=\mu^f_j})$ is always positive in all dimensions. See proof in Appendix \ref{proof_claim_2}. 
\end{claim}
According to Eq. (\ref{eq:lemma2}) introduced in Lemma \ref{lemma_1} and Claim \ref{claim2}, we conclude that $\text{Cov}(\textbf{g}_i,\textbf{g}_j)$ and $\sigma_{x_i^f,x_j^f}$ are positively and monotonically correlated. Therefore, $\text{Cov}(\textbf{g}_i,\textbf{g}_j)$ and $I(\textbf{x}_i; \textbf{x}_j)$ are also positively and monotonically correlated, which completes the proof.
\end{proof}

Our theoretical findings may also be applicable to Federated \textit{Supervised} Domain Generalization, which could form the subject of future research as expressed in Appendix \ref{proof:supervised} and \cite{molahasani2024theoretical}. We further investigate the relationship between the distribution of the domains and their corresponding gradients through the following Corollary. 
\setcounter{theorem}{0}
\begin{corollary}\label{corollary1}
Given the assumptions stated, for two distinct domains characterized by random variables $\textbf{\emph{x}}_i$ and $\textbf{\emph{x}}_j$ from the distribution family $\mathcal{P}$, as 
$I(\textbf{\emph{x}}_i,\textbf{\emph{x}}_j)$ decreases,
then the variance of the difference of the corresponding gradients, $\emph{\text{Var}}(\textbf{\emph{g}}_i - \textbf{\emph{g}}_j)$, increases. The full proof is presented in Appendix \ref{Ap:rem1}.
\end{corollary}

\subsection{FedGaLA}
\label{secsec:FSSDG}

The analysis above establishes a link between gradient alignment and domain shift, forming the basis for our proposed FedGaLA framework. 
Figure \ref{fig:overview} illustrates an overview of our framework for addressing the newly proposed problem setup (Definition \ref{def1}). 
FedGaLA is remotely inspired by prior works that demonstrate improved generalization in centralized (non-federated) learning through gradient alignment \cite{mansilla2021domain,shi2022gradient,rame2022fishr}. However, our framework extends this notion by integrating gradient alignment into the field of unsupervised and federated learning, and assumes, unlike \cite{mansilla2021domain,shi2022gradient,rame2022fishr}, that the distribution of data from different clients is not known. 
Our core idea includes (\textit{i}) enabling clients to learn domain-invariant representations at the client level through local gradient alignment, and (\textit{ii}) adjusting the aggregation weights at the server level using global gradient alignment.

\begin{center}
\begin{minipage}[b]{0.45\textwidth}
\begin{algorithm}[H]
\small
   \caption{FedGaLA}
   \label{alg:UFDG}
   \begin{algorithmic}[1]
      \State \textbf{Input:} data $\mathcal{D}_i$, initialization $\Theta^0$
      \State \textbf{Output:} $\Theta^{T}$
      \For{$t  \hspace{0.2cm} \text{from}\hspace{0.2cm} 1 \hspace{0.2cm}\text{to} \hspace{0.2cm} T$}
         \State \textbf{Server:}
         \State \hspace{\algorithmicindent} Calculate client updates: $\textbf{g}_i^{(t)} \gets \Theta^{(t)}_i - \Theta^{(t-1)}$
         \State \hspace{\algorithmicindent} Initialize global update: $\textbf{g}^{(t+1)} \gets \text{FedAVG}(\textbf{g}_i^{(t)})$
         \For{$j  \hspace{0.2cm} \text{from}\hspace{0.2cm} 1 \hspace{0.2cm}\text{to} \hspace{0.2cm}iter$}
            \State \hspace{\algorithmicindent} Calculate weights: $w_{i} \gets \frac{\text{Cosine}(\textbf{g}_i^{(t)}, \textbf{g}^{(t+1)}) + 1}{2}$
            \State \hspace{\algorithmicindent} Normalize weights: $w_{i} \gets \frac{w_{i}}{\sum w_{i}} $
            \State \hspace{\algorithmicindent} Aggregate updates: $\hat{\textbf{g}}^{(t+1)} \gets \sum_{i=1}^K w_{i}\textbf{g}_i^{(t)}$
         \EndFor
         \State \hspace{\algorithmicindent} Update global model: $\Theta^{(t+1)} \leftarrow \Theta^{(t)}+\hat{\textbf{g}}^{(t+1)}$
         \State \hspace{\algorithmicindent} Communicate: $\Theta_i \gets \Theta^{(t+1)}$
         \State \textbf{Client:}
         \For{$j  \hspace{0.2cm} \text{from}\hspace{0.2cm} 1 \hspace{0.2cm}\text{to} \hspace{0.2cm} N_{\text{batch}}$}
            \State Compute batch gradient: $\textbf{g}_{i,j}^{(t)} \gets \nabla l_i(x_{i,j},\theta_i^{(t)})$
             \For{$l  \hspace{0.2cm} \text{from}\hspace{0.2cm} 1 \hspace{0.2cm}\text{to} \hspace{0.2cm}L$}
             \State  Estimate reference: $\hat{\textbf{g}}_{est}^{(l,t)}=\theta^{(l,t)}-\theta^{(l,t-1)}$
                 
            \If{$\text{Cosine}(\textbf{g}_{i,j}^{(l,t)},\hat{\textbf{g}}_{\text{est}}^{(l,t)}) > \tau$}
               \State Update weights: 
               \Statex \hspace{2cm}  $\theta_i^{(l,t)} \gets \theta_i^{(l,t-1)} - \eta \cdot\nabla l_i^{(l,t)}(x_{i,j},\theta_i^{(l,t)})$
            \EndIf
            \EndFor
            \EndFor
         
         \State \hspace{\algorithmicindent} Communicate: $S \gets \Theta_i$
      \EndFor
   \end{algorithmic}
\end{algorithm}
\end{minipage}
\end{center}

At each communication round, clients are initialized with the global model. Subsequently, each client updates its parameters using SSL for $E$ epochs based on the local data through local gradient alignment and sends these updates back to the server. Finally, the server employs the global gradient alignment technique to perform aggregation. This procedure is repeated for $T$ communications rounds to determine the global model. The remaining part of this section provides details on local and global gradient alignment, and the complete framework is outlined in Algorithm \ref{alg:UFDG}. 

\noindent \textbf{Local gradient alignment.}
\label{secsec:GA}
Our method performs layer-wise local gradient alignment using a reference gradient. This reference is derived from the $l^\text{th}$ layer of the global model's parameter updates between the current and the previous communication round. Suppose $\Theta= \{\theta^{(l)} \}_{l=1}^{L}$ indicates the parameters of the global model, where $\theta^{(l)}$ represents the parameters of the $l^\text{th}$ layer. The reference gradient for the $l^{th}$ layer is computed as $\hat{\textbf{g}}_{est}^{(l,t)}=\theta^{(l,t)}-\theta^{(l,t-1)}$, where $\theta^{(l,t)}$ and $\theta^{(l,t-1)}$ are the parameters of the $l^\text{th}$ layer of the global model at rounds $t$ and $t-1$, respectively. Then, $\hat{\textbf{g}}_{est}^{(l,t)}$ is locally used to determine whether the gradient of each layer obtained during training (e.g., via SGD) is aligned with the reference. The cosine similarity between the batch gradient and the reference for each layer $l$ at round $t$ is computed as $cos(\phi)^{(l,t)} = \frac{\langle \textbf{g}_{i,k}^{(l,t)}, \textbf{g}_{\text{est}}^{(l,t)} \rangle}{\|\textbf{g}_{i,k}^{(l,t)}\| \cdot \|\textbf{g}_{\text{est}}^{(l,t)}\|}$, where $\textbf{g}_{i,k}^{(l,t)}$ is the gradient of $k^\text{th}$ batch of the $i^\text{th}$ client at layer $l$ and round $t$. Finally, batch gradients whose similarity with $\hat{\textbf{g}}_{est}^{(l,t)}$ are less than a user-defined threshold $\tau$ are discarded during the update process. This prevents clients from learning domain-specific features by disregarding local gradients that are not aligned with the global model.
The rationale for discarding unaligned gradients instead of applying soft weighting is that when a gradient vector is unaligned with  $\hat{\textbf{g}}_{est}$, scaling does not change this alignment, as the cosine of the angle between them is independent of scale.
To establish a theoretical basis for the proposed local alignment, we introduce the following.
\setcounter{theorem}{0}
\begin{proposition}\label{proposition}
Given two sets of gradient vectors $\textbf{\emph{g}}_{i}$ and $\textbf{\emph{g}}_{j}$, by removing the $K^{th}$ vector in  $\textbf{\emph{g}}_{j}$ where $cos(\textbf{\emph{g}}_{j,K},\textbf{\emph{g}}_{est})<0$, the covariance of two sets increases. For more details on this Proposition and its proof, see Appendix \ref{prop_proof}. 
\end{proposition}

Theorem \ref{theorem} highlighted how an increase in domain shift between clients $C_i$ and $C_j$ correlates with a decrease in the covariance of their gradient vectors, $\textbf{g}_i$ and $\textbf{g}_j$. Therefore, gradient covariance can be used as an indicator of domain shift. Proposition \ref{proposition} complements this by showing that by selectively removing gradients from $\textbf{g}_j$ that have a negative cosine similarity with an estimated target direction, $\textbf{g}_{est}$, we can effectively increase the covariance of the gradient sets, thus potentially counteracting the effects of domain shift.

\noindent \textbf{Global gradient alignment.}
\label{secsec:AgbA}
To aggregate the local (client) models at the server, the locally measured gradients that closely match the average gradient across all clients are assigned greater weights. This soft weighting process operates as follows. Once the server receives the local models, it first obtains the local update $\hat{\textbf{g}}_i^{(t)}=\Theta_i^{(t)}-\Theta^{(t-1)}$. Then, the initial global update  $\hat{\textbf{g}}^{(t+1)}$ is calculated by averaging all $\hat{\textbf{g}}_i^{(t)}$. Subsequently, the weight of each client is computed using $w_{i} = \frac{cos(\hat{\textbf{g}}_i^{(t)}, \hat{\textbf{g}}^{(t+1)}) + 1}{2}$, where $w_i$ is the weight for the $i^\text{th}$ client and $\hat{\textbf{g}}_i^{(t)}$ is the gradient of the $i^\text{th}$ client at round $t$. This weight reflects the degree of alignment between each client's update and the global model's update direction. To ensure these weights are properly normalized using $w_i = {w_i}/{\sum_{k=1}^K w_k}$. The normalization of weights allows for the proportional contribution of each client's update. 

Finally, the normalized weights are used to perform aggregation. Each client's model update is scaled by its respective weight, and these weighted updates are then aggregated to compute the weighted average update. The global model at round $t+1$ is updated based on $\hat{\textbf{g}}^{(t+1)} = \sum_{i=1}^K w_i \hat{\textbf{g}}_i^{(t)}$. This aggregation step is repeated three times, refining the weights with each iteration. This ensures the global model update is significantly influenced by clients whose updates align closely with the global learning objective. After completing the aggregation process, the global model is further updated based on $\Theta^{(t+1)} \leftarrow \Theta^{(t)}+\hat{\textbf{g}}^{(t+1)}$.
The rationale for using different alignment strategies at the client and server levels stems from the inherent differences between local training and global aggregation. 
At the client level, we manage batch gradients, allowing us to specifically discard the unaligned ones without significant information loss. 
However, discarding gradients at the server-level corresponds to the deletion of entire clients, which can adversely affect the outcome.

\section{Experiment setup}
\label{sec:exp}

\textbf{Datasets.}
\label{secsec:data}
To evaluate the effectiveness of our proposed method, we conduct experiments across four commonly used benchmarks for domain generalization. They include: \textbf{PACS} \cite{li2017deeper}, which consists of 9,991 images from four domains: `Photo', `Art-painting', `Cartoon', and `Sketch', across seven classes; \textbf{Office-Home} \cite{venkateswara2017deep}, which consists of 15,588 images from four domains: `Art', `Clipart', `Product', and `Real-world', across 65 classes; \textbf{TerraInc} \cite{beery2018recognition}, which includes 24,788 images from four domains `Location 38', `Location 43', `Location 46', and `Location 100', across nine classes; \textbf{DomainNet} \cite{peng2019moment}, which consists of 569,010 images in six domains: `Clipart', `Infograph', `Painting', `Quickdraw', `Real', and `Sketch', covering 345 classes. 
Following \cite{zhang2022towards}, for the DomainNet dataset, we select the following classes: zigzag, tiger, tornado, flower, giraffe, toaster, hexagon, watermelon, grass, hamburger, blueberry, violin, fish, sun, broccoli, Eiffel tower, horse, train, bird, and bee \cite{zhang2022towards}, resulting in a total of 38556 samples. In all experiments using this dataset, three domains are used for training (`Painting', `Real', and `Sketch') and the other three domains (`Clipart',  `Infographics', and `Quickdraw') are used for testing.

\noindent \textbf{Evaluation.}
We use the leave-one-domain-out setting used in prior works \cite{zhang2023federated,zhang2022towards, gulrajani2021in}. This involves selecting one domain as the target, training the model on the rest of the domains, and then testing the model's performance on the selected target domain. Linear evaluation, a common feature evaluation approach, is utilized to evaluate the quality of learned representations \cite{feng2019self,zhang2017split,kolesnikov2019revisiting}. For linear evaluation, following \cite{van2020towards, zhuang2022divergence}, we utilize 10\% and 30\% of the target data to train the linear classifier and evaluate the remaining 90\% and 70\% of the data, respectively. 

\begin{table*}[t]
\setlength{\tabcolsep}{7pt}
\centering
\resizebox{0.95\linewidth}{!}{
\begin{tabular}{l|c|ccccccccc}
\toprule 
\multirow{2}{*}{\textbf{Model}} & \textbf{Labeled} & \multicolumn{5}{c|}{\textit{\textbf{PACS}}} & \multicolumn{4}{c}{\textit{\textbf{DomainNet}}} \\ 
& \textbf{ratio} & P & A & C & S & \multicolumn{1}{c|}{Ave.} & C & I & Q & Ave. \\ \midrule \midrule
\multicolumn{1}{l|}{FedEMA} &\multirow{6}{*}{10\%} & 50.0(0.7) & 29.5(1.9) & 42.4(2.3) & 45.6(0.16) & \multicolumn{1}{c|}{41.9} & 38.6(1.1) & 13.7(0.5) & 45.5(1.6) & 32.4 \\ 
\multicolumn{1}{l|}{FedBYOL} & &52.1(1.1) & 31.8(1.1) & 45.4(2.2) & 47.4(1.9) & \multicolumn{1}{c|}{44.2} & 38.1(0.5) & 14.1(0.4) & 53.6(2.9) & 31.8 \\ 
\multicolumn{1}{l|}{FedMoCo} & &58.5(2.2) & 35.7(9.6) & 37.7(12.9) & 36.6(8.2) & \multicolumn{1}{c|}{42.1} & 30.5(0.6) & 10.9(2.1) & 46.4(0.8) & 27.2 \\ 
\multicolumn{1}{l|}{FedSimSiam} & & 46.2(1.1) & 28.6(1.0) & 46.7(0.6) & 37.6(1.3) & \multicolumn{1}{c|}{39.8} & 44.8(1.5) & 12.2(0.3) & 40.3(2.4) & 36.9 \\ 
\multicolumn{1}{l|}{FedSimCLR} & &64.2(1.2) & 41.9(1.5) & 58.4(1.3) & 70.1(1.2) & \multicolumn{1}{c|}{58.6} & 45.2(0.4) & 13.7(0.3) & 59.7(0.7) & 39.5 \\ 
\multicolumn{1}{l|}{  FedGaLA (ours)} & &\textbf{64.7(1.9)} & \textbf{44.2(1.2)} & \textbf{60.5(2.2)} & \textbf{70.5(1.3)} & \multicolumn{1}{c|}{\textbf{60.0}} & \textbf{47.6(0.9)} & \textbf{14.2(0.5)} & \textbf{61.4(0.3)} & \textbf{41.1} \\ 
\midrule
\multicolumn{1}{l|}{FedEMA} &  \multirow{6}{*}{30\%} & 53.5(0.3) & 33.9(2.8) & 48.3(2.4)  & 45.3(2.3) & \multicolumn{1}{c|}{45.2} & 43.5(0.8) &20.0(1.5) & 49.0(2.8) & 37.5 \\ 
\multicolumn{1}{l|}{FedBYOL} & & 55.1(1.2) & 35.3(1.5) & 48.3(1.5) & 48.9(0.5) & \multicolumn{1}{c|}{46.9} & 44.6(1.2)  &\textbf{20.5(1.1)} & 50.4(0.9) & 38.5 \\ 
\multicolumn{1}{l|}{FedSimSiam} & &  58.5(2.2) & 35.7(9.6) & 37.7(12.9) & 36.6(8.2) & \multicolumn{1}{c|}{42.1} & 35.7(1.1) & 14.9(3.9) & 44.6(3.5) & 31.8 \\ 
\multicolumn{1}{l|}{FedMoCo} & & 47.3(1.6) & 30.2(3.4) & 47.4(0.7) & 34.4(1.7) & \multicolumn{1}{c|}{39.9} & 51.5(1.8)  & 18.2(0.1) & 59.9(3.6)& 43.2 \\  
\multicolumn{1}{l|}{FedSimCLR} & & 69.8(1.1) & 46.4(2.1) & 63.9(1.6) & 73.4(3.0) & \multicolumn{1}{c|}{63.3} & 51.7(1.0)&  16.3(0.2) & 66.5(0.9) & 44.8 \\  
\multicolumn{1}{l|}{ FedGaLA (ours)} & & \textbf{71.1(2.0)} & \textbf{46.8(2.1)} & \textbf{65.7(1.6)} & \textbf{74.5(1.1)} & \multicolumn{1}{c|}{\textbf{64.6}} & \textbf{ 52.4(0.7)} & 16.2(0.6)& \textbf{68.8(0.6)}& \textbf{45.8} \\ \bottomrule 
\end{tabular}%
}
\vspace{1mm}
\centering
\resizebox{0.95\linewidth}{!}{
\begin{tabular}{l|c|cccccccccc}
\toprule
 \multirow{2}{*}{\textbf{Model}} & \textbf{Labeled}  & \multicolumn{5}{c|}{\textit{\textbf{Office-Home}}} & \multicolumn{5}{c}{\textit{\textbf{TerraInc}}} \\
 & \textbf{ratio} & \multicolumn{1}{c}{A} & C & P & R & \multicolumn{1}{c|}{Ave.} & L38 & L43 & L46 & L100 & Ave. \\ 
\midrule \midrule
\multicolumn{1}{l|}{FedEMA} & \multirow{6}{*}{10\%} & 6.7(0.5) & 12.5(0.4) &20.8(0.9)   & 14.1(0.8) & \multicolumn{1}{c|}{13.5} & 62.1(0.4) & 43.5(3.0) & 43.4(0.4) & 63.9(0.7) & 53.2 \\ 
\multicolumn{1}{l|}{FedBYOL} & & 7.4(0.3) & 12.9(0.5) & 20.9(1.1) & 13.8(0.1) & \multicolumn{1}{c|}{13.8} & 63.4(0.2) & 43.3(0.5) & 44.3(0.3) & 66.3(2.4) & 54.3 \\
\multicolumn{1}{l|}{FedSimSiam} & & 8.5(0.5) & 19.8(0.8) & 28.2(0.8) & 16.2(0.6) & \multicolumn{1}{c|}{18.9} & 54.0(6.8) & 36.0(1.5) & 40.9(3.7) & 60.8(1.6) & 47.9 \\ 
\multicolumn{1}{l|}{FedMoCo} & & \textbf{10.8(0.4)} & 9.4(0.3) & 12.4(0.9) & 10.1(0.7) & \multicolumn{1}{c|}{10.7} & 50.0(2.4) & 33.8(0.3) & 29.7(2.3) & 60.2(0.1) & 45.7 \\
\multicolumn{1}{l|}{FedSimCLR} & & 8.9(0.4) & 24.3(0.3) & 35.2(1.2) & 20.0(0.2) & \multicolumn{1}{c|}{22.0} & 62.8(0.2) & 45.8(1.1) & 42.9(1.8) & 68.8(0.3) & 55.1 \\ 
\multicolumn{1}{l|}{FedGaLA (ours)} & & 8.9(0.4) & \textbf{25.3(0.6)} & \textbf{36.6(0.3)} & \textbf{21.2(0.5)} & \multicolumn{1}{c|}{\textbf{23.0}} & \textbf{63.6(0.1)} & \textbf{47.6(1.4)} & \textbf{43.9(2.9)} & \textbf{71.7(0.9)} & \textbf{56.7} \\ \midrule

\multicolumn{1}{l|}{FedEMA} & \multirow{6}{*}{30\%}  & 10.0(0.3) & 17.3(0.7) & 28.2(1.1) & 17.8(0.3) & \multicolumn{1}{c|}{18.2} & 62.7(0.6) & 47.1(0.3) & 46.1(0.4) & 67.6(2.0) & 55.9 \\ 
\multicolumn{1}{l|}{FedBYOL} &  & 10.4(0.3) & 17.4(1.2) & 29.4(1.3) & 17.6(0.6) & \multicolumn{1}{c|}{18.7} & 63.6(0.9) & 46.7(1.3) & 46.1(0.2) & 68.8(2.5) & 56.3 \\  
\multicolumn{1}{l|}{FedSimSiam} & & 12.7(1.1) & 24.1(1.9) & 36.5(0.7) & 21.6(1.1) & \multicolumn{1}{c|}{23.7} & 56.8(4.6) & 35.7(2.0) & 38.9(4.6) & 61.4(1.4) & 48.2 \\  
\multicolumn{1}{l|}{FedMoCo} & & 15.2(0.6) & 11.2(0.9) & 15.8(0.7) & 11.5(1.6) & \multicolumn{1}{c|}{13.4} & 60.7(0.1) & 28.3(3.3) & 31.7(2.3) & 60.0(0.6) & 45.2 \\
\multicolumn{1}{l|}{FedSimCLR} &  & \textbf{13.6(1.3)} & \textbf{35.3(0.7)} & 47.6(0.9) & 26.6(1.0) & \multicolumn{1}{c|}{30.7} & 65.8(0.3) & \textbf{54.5(1.2)} & 49.0(0.4) & 71.0(1.6) & 60.1 \\  
\multicolumn{1}{l|}{ FedGaLA (ours)} & &  \textbf{13.6(0.6)} & 35.1(0.2) & \textbf{48.0(1.3)} & \textbf{27.0(0.7)} & \multicolumn{1}{c|}{\textbf{30.9}} & \textbf{66.2(0.5)} & 52.4(1.7) & \textbf{51.7(0.8)} & \textbf{74.6(1.3)} & \textbf{61.3} \\ \bottomrule
\end{tabular}%
}
\caption{Results of linear eval. on PACS,  DomainNet, Office-Home, and TerraInc datasets.}
\label{tab:results}
\end{table*}

\noindent \textbf{Baselines.} 
To evaluate our method, we take a two-pronged approach: (\textbf{1}) We adapt several popular SSL approaches to the federated domain generalization task, denoting them as FedSimCLR, FedMoCo, FedBYOL, and FedSimSiam.
To this end, we employed SimCLR \cite{chen2020simple}, MoCo \cite{he2020momentum}, BYOL \cite{grill2020bootstrap}, and SimSiam \cite{chen2021exploring} in a federated setup. Each of these models consists of two encoders. BYOL and MoCo utilize exponential moving averages to update one of the encoders (target network) using the other encoder (online network). In contrast, SimCLR and SimSiam share weights between their two encoders. Additionally, SimCLR and MoCo, being contrastive-based SSL models, leverage negative samples in the learning process. Conversely, BYOL and SimSiam, which do not rely on negative samples, include a predictor on top of the online encoder to enhance learning from unlabeled samples.
We then train each client locally using the respective SSL method. Next, we aggregate the trained encoders at the server using FedAVG \cite{mcmahan2017communication}. For BYOL and SimSiam, we follow the procedure in \cite{zhuang2021collaborative} and apply FedAVG on the online encoder and projector. We also adapt FedEMA \cite{zhuang2022divergence}, which is a commonly used method originally developed for federated unsupervised learning. FedEMA integrates BYOL as an SSL technique into its structure.

 It is important to note that we avoid direct comparisons with FedDG \cite{liu2021feddg}, FedSR \cite{nguyen2022fedsr}, FedADG 
\cite{zhang2021federated}, FedIIR \cite{guo2023out}, and FedDG-GA \cite{zhang2023federatedGA} since unlike our method, they employ the label information in their solutions. 
(\textbf{2}) Given the absence of prior research specifically addressing the problem of \textit{federated} unsupervised domain generalization, we also compare FedGaLA to established \textit{centralized unsupervised domain generalization} methods on the PACS and DomainNet datasets. We could not identify unsupervised domain generalization methods for Office-Home and TerraInc datasets. This comparison includes the following solutions: SimCLR \cite{chen2020simple}, MoCo \cite{he2020momentum}, BYOL \cite{grill2020bootstrap}, AdCo \cite{hu2021adco}, and DARLING \cite{zhang2022towards}. It is important to note that these methods are implemented in a non-federated environment and do not incorporate any data privacy constraints. All the results for these models are reported from \cite{zhang2022towards}.

\noindent \textbf{Implementation details.} 
We use SimCLR as the SSL module in FedGaLA due to its performance on domain generalization problems as previously shown \cite{zhang2022towards}. Following\cite{zhang2022towards}, ResNet-18 \cite{he2016deep} is employed as the encoder network architecture for all experiments, which we train from scratch. We present details regarding data augmentations, projector architecture, and encoder hyperparameters in Appendix \ref{app:imp}. Following \cite{feng2019self,zhang2017split}, we first learn a representation by FedGaLA and the baseline models for 100 communication rounds with 7 local epochs. 
Next, we freeze the backbone model and train a liner classier for 100 epochs to perform prediction on the target domain. Appendix \ref{app:imp} presents all the hyperparameters used for FedGaLA. For FedEMA, we use the hyperparameters reported in \cite{zhuang2022divergence}. All experiments were implemented using PyTorch and trained on 8 NVIDIA GeForce RTX 3090 GPUs. For each experiment, we train the models three times with random initialization seeds and report the average.

\noindent  {\textbf{Cross-silo vs. cross-device federated learning.} Two frameworks can generally be considered in federated learning: cross-silo and cross-device. In cross-silo, the number of clients $K$ is small, and each client is capable of retaining its local model and state across training rounds. In contrast, cross-device federated learning works under the assumption of having millions of stateless clients, where only a limited number of clients participate in training during each communication round. However, due to limitations in experimental settings, the majority of studies typically conduct experiments involving at most hundreds of clients \cite{Wang2020Federated,zhuang2022divergence}. In this study, given the limited number of domains FedGaLA works within the cross-silo framework. Moreover, each client is equipped with the necessary resources to participate in the collaborative learning process. Additionally, we ensure that every client updates its local model with a sufficient number of epochs.

\begin{figure*}[t]
    \centering
    \begin{subfigure}{0.22\linewidth}
     \includegraphics[width=\linewidth]{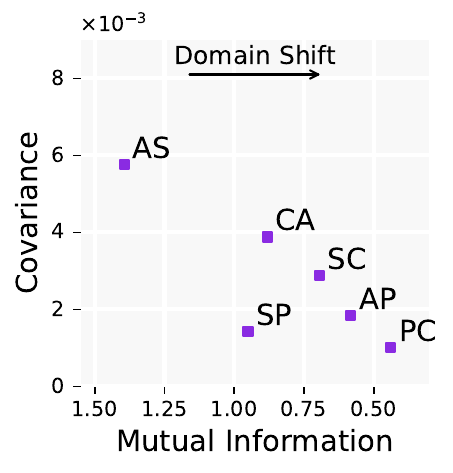}
     \caption{} 
        \label{fig:cov}
    \end{subfigure}
     \hfill
    \begin{subfigure}{0.22\linewidth}
        \includegraphics[width=\linewidth]{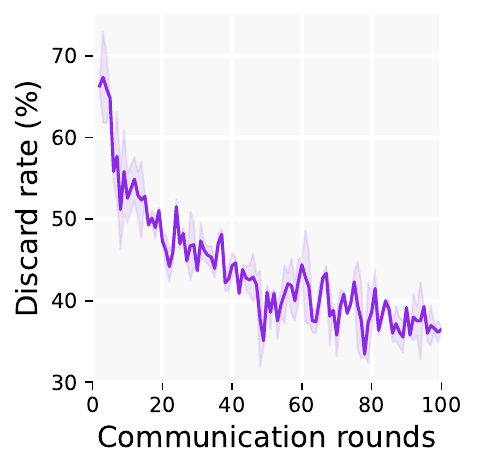}
        \caption{} 
        \label{fig:communication}
    \end{subfigure}
     \hfill
    \begin{subfigure}{0.22\linewidth} 
        \centering
        \includegraphics[width=0.97\linewidth]{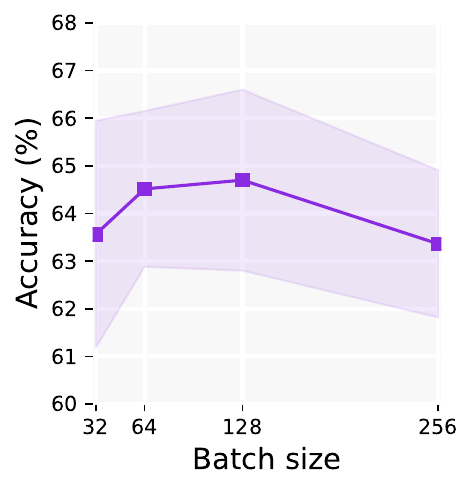}
        \caption{}
        \label{fig:batchh}
    \end{subfigure}
     \hfill
    \begin{subfigure}{0.22\linewidth}
        \includegraphics[width=1.1\linewidth]{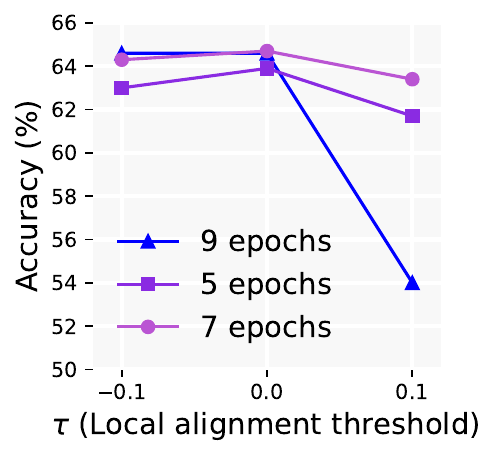}
        \caption{} 
        \label{fig:TouE}
    \end{subfigure}

    \caption{(a) Covariance under domain shift across various domains of the PACS dataset. (b) Average ratio of discarded local gradients over 100 communication rounds ($\tau=0$ and $E=1$). (c) Impact of various batch sizes. (d) Impact of $\tau$ on performance across different number of local epochs. Results in (b-d) are reported for the domain $P$ of the PACS dataset, with a 10\% label ratio.}
    \label{fig:parameter1}
\end{figure*}

\begin{table}[t]
\centering
\setlength{\tabcolsep}{2pt}
\resizebox{1\columnwidth}{!}{
\begin{tabular}{l|ccccccccc}
\toprule 
 \multirow{2}{*}{Model} & \multicolumn{5}{c|}{\textit{PACS}} & \multicolumn{4}{c}{\textit{DomainNet}} \\ 
 & P & A & C & S & \multicolumn{1}{c|}{Ave.} & C & I & Q & Ave. \\ \midrule \midrule
BYOL & 27.0 & 25.9 & 21.0 & 19.7 & \multicolumn{1}{c|}{23.4} & 14.6 & 8.7 & 5.9 & 9.7 \\
MoCo & 44.2 & 25.9 & 33.5 & 25.0 & \multicolumn{1}{c|}{32.1} & 32.5 & 18.5 & 8.1 & 19.7 \\
SimCLR & 54.7 & 37.7 & 46.0 & 28.3 & \multicolumn{1}{c|}{41.6} & 37.1 & 19.9 & 12.3 & 23.1 \\
AdCo & 46.5 & 30.2 & 31.5 & 22.9 & \multicolumn{1}{c|}{32.8} & 32.3 & 17.9 & 11.6 & 20.6 \\
DARLING & 53.4 & 39.9 & 46.4 & 30.2 & \multicolumn{1}{c|}{42.5} & 35.2 & 20.9 & 15.7 & 23.9 \\ 
FedGaLA (ours) & \textbf{64.7} & \textbf{44.2} & \textbf{60.5} & \textbf{70.5} & \multicolumn{1}{c|}{\textbf{60.0}} & \textbf{47.6} & \textbf{14.2} & \textbf{61.4} & \textbf{41.1} \\ \bottomrule
\end{tabular}
}
\caption{Comparison of linear eval. (10\%) results with non-federated (centralized) domain generalization methods on PACS and DomainNet datasets.}
\label{tab:pacs_domainnet_cent}
\end{table}

\section{Results}
\label{sec:results}

In this section, we first present the performance evaluation of FedGaLA across various datasets, comparing it against baseline methods to demonstrate its effectiveness. This is followed by ablation studies and sensitivity analyses, which provide detailed insights into the contributions of individual components and the robustness of our proposed framework.

\subsection{Performance}
\textbf{Federated unsupervised domain generalization.} We report the accuracy rates of FedGaLA and baseline models on the four datasets. As shown in Table \ref{tab:results}, FedGaLA consistently outperforms all baselines across all four datasets, with the exception of the `Art-painting’ domain in Office-Home, for the 10\% data regime. When 30\% of the data are used, our method still generally outperforms the baseline models, although, for some of the domains, the baseline solutions produce slightly better results. This observation is expected given that with the introduction of more domain-specific training data, the need for domain generalization declines, and thus, methods that are not explicitly designed for domain generalization can produce competitive results. 
Across the four datasets, we observe that among the baseline models, FedSimCLR achieves better results compared to FedSimSiam, FedBYOL, and FedMoCo. This finding is consistent with \cite{zhang2022towards} where it was demonstrated that SimCLR provides a better foundation for domain generalization versus other SSL methods, albeit in a non-federated setup. 

\noindent \textbf{Centralized unsupervised domain generalization.}
We compare the performance of FedGaLA with centralized (non-federated) methods for PACS and DomainNet datasets, where the baselines are trained on the entire dataset consisting of all the domains. The results presented in Table \ref{tab:pacs_domainnet_cent} show that FedGaLA outperforms centralized methods by large margins.
This is an expected observation as prior works have shown that federation can boost domain generalization \cite{nguyen2022fedsr,arasteh2023mind}.

\begin{table}[t]
\begin{center}
\begin{adjustbox}{width=\columnwidth,center}
\begin{small}
\begin{tabular}{lccccc}
\toprule
Model &  P & A & C & S & Ave.\\
\midrule \midrule
FedMoCo       & 46.2(1.1) & 28.6(1.0) & 46.7(0.6) & 37.6(1.3) &  39.8\\
FedGaLA w/ MoCo & \textbf{46.5(0.3) }& \textbf{28.9(0.4)} & \textbf{47.9(0.3)} & \textbf{39.6(2.4)} &\textbf{40.7} \\                    
 \cmidrule{1-6}
FedBYOL    & 52.1(1.1) & \textbf{31.8(1.1)} & 45.4(2.2) & 47.4(1.9) & 44.2 \\
FedGaLA w/ BYOL & \textbf{52.8(0.6)} & 31.7(0.5) & \textbf{46.3(3.1)} & \textbf{47.6(1.2)} & \textbf{44.6}\\
 \cmidrule{1-6}
FedSimCLR       & 64.2(1.2) & 41.9(1.5) & 58.4(1.3) & 70.1(1.2) & 58.6 \\
FedGaLA w/ SimCLR  &\textbf{64.7(1.9)} & \textbf{44.2(1.2)} & \textbf{60.5(2.2)} & \textbf{70.5(1.3)} & \textbf{60.0} \\

\bottomrule
\end{tabular}
\end{small}
\end{adjustbox}
\end{center}
\caption{The performance of FedGaLA when employing different SSL methods.}
\label{Tab:ssl}
\end{table}

\begin{table}[t]

\begin{center}
\resizebox{1\columnwidth}{!}{
\begin{tabular}{lccccc}
\toprule
Model &  P & A & C & S & Ave.\\
\midrule \midrule
FedGaLA     &  \textbf{64.7(1.9)} & \textbf{44.2(1.2)} & \textbf{60.5(2.2)} & \textbf{70.5(1.3)} & \textbf{60.0}\\ 
w/o GA      & \textbf{64.7(0.4)} & 42.6(1.2) & 58.1(0.6) & 69.9(1.1) & 58.8\\
w/o LA       & 63.7(1.5) & 41.6(1.4) & 59.8(1.5) & 68.9(1.1) &  58.5\\
w/o GA \& LA  &  64.2(1.2) & 41.9(1.5) & 58.4(1.3) & 70.1(1.2) & 58.6 \\
\bottomrule
\end{tabular}
}
\end{center}
\caption{Ablation study on PACS (GA: global alignment; LA: local alignment).}
\label{Tab:ablation}
\end{table}

\begin{figure*}[t] 
    \centering
        \begin{subfigure}{0.23\linewidth} 
        \centering
        \includegraphics[width=0.99\linewidth]{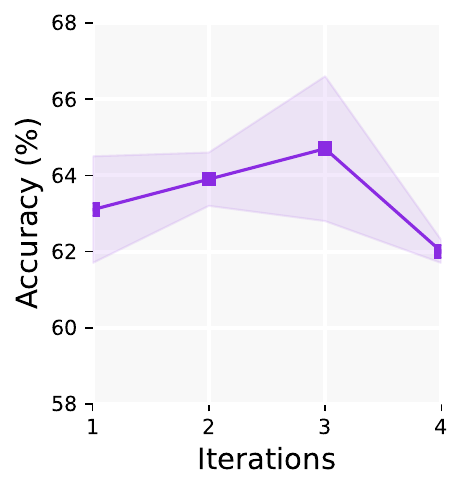}
        \caption{}
        \label{fig:iteration}
    \end{subfigure}
     \hfill
         \begin{subfigure}{0.23\linewidth} 
        \centering
        \includegraphics[width=0.97\linewidth]{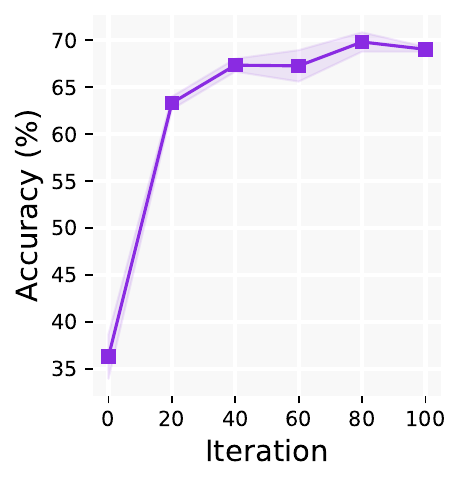}
        \caption{}
        \label{fig:stability}
    \end{subfigure}
     \hfill
    \centering
             \begin{subfigure}{0.22\linewidth}
        \includegraphics[width=1.04\linewidth]{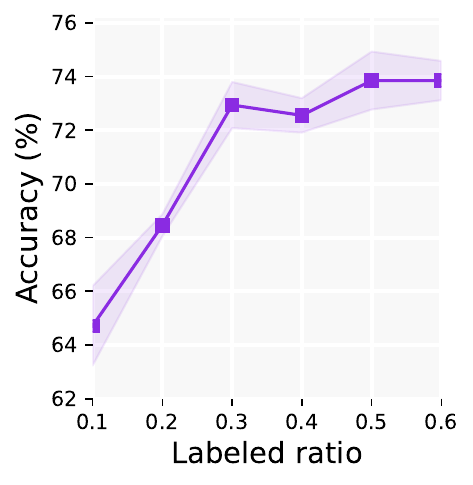}
        \caption{} 
        \label{fig:labeled}
    \end{subfigure}
     \centering
    \hfill 
    \begin{subfigure}{0.23\linewidth}
        \includegraphics[width=1.01\linewidth]{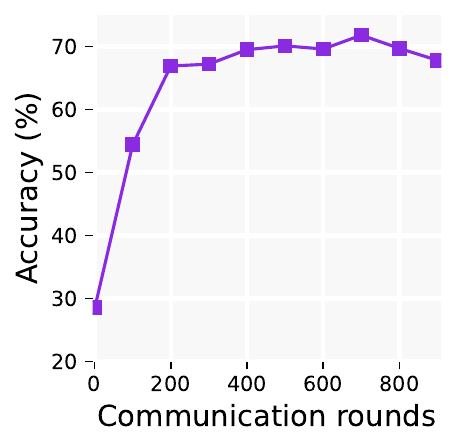}
        \caption{} 
        \label{fig:T}
    \end{subfigure}
    \hfill 
     \centering

    \caption{ (a) Impact of the number of iterations for global alignment. (b) Stability of the training. (c) Effect of different labeled data ratios on linear evaluation performance. (d) Performance of the model in different communication rounds. Results are reported for domain $P$ of PACS dataset with $10\%$ label ratio, except for (c) where the label ratio is changing.}
    \label{fig:sensi}
\end{figure*}

\noindent \textbf{FedGaLA with other SSL techniques.}  
Recall that we used SimCLR in our federated framework. To further explore whether FedGaLA can improve the performance of other federated SSL methods, 
we perform additional experiments in which we employ MoCo and BYOL in FedGaLA instead of SimCLR. We evaluate the performance on the PACS dataset, as presented in Table \ref{Tab:ssl}. The results show that FedGaLA consistently improves the performance of all federated SSL frameworks, demonstrating its ability to adapt to different SSL methodologies while boosting domain generalization.

\subsection{Empirical verification of Theorem \ref{theorem}}
\noindent \textbf{Gradient misalignment due to domain shift.} In Figure \ref{fig:cov} we illustrate the amount of measured covariance for different domains in PACS versus the amount of domain shift between each pair measured through Mutual Information. We observe that, except for a single outlier, the trend follows our prediction based on Theorem \ref{theorem}.

\subsection{Ablation studies and component analysis}
\label{secsec:AS}
\noindent \textbf{Ablation studies.}
Here we examine the effectiveness of the local and global gradient alignment components individually on the final performance of FedGaLA. To this end, we systematically remove each of these components. As shown in Table \ref{Tab:ablation}, each component plays an important role in the overall performance. It is noteworthy to mention that FedGaLA essentially becomes the FedSimCLR baseline by removing both global and local alignments.  

\noindent \textbf{Ratio of discarded local gradients.} Figure \ref{fig:communication} demonstrates the ratio of local gradients discarded due to local gradient alignment during training versus communication rounds. For this experiment, the local models are trained for 1 epoch at each communication round. As observed, the ratio of discarded local gradients decreases from approximately 68\% in the early communication rounds to 37\% by round 100. This trend indicates that as the number of communication rounds increases, the local gradient directions become more aligned with the global model, suggesting that with increasing the number of communications, local models learn more domain-invariant features.

\begin{table}[t]
\centering
\begin{adjustbox}{width=0.7\columnwidth,center}
\begin{tabular}{l|cccc}
\toprule
\multirow{1}{*}{ Model} & \multicolumn{4}{c}{\textit{Communication Frequency ($E$)}} \\  
 & $E=1$ & $E=5$ & $E=7$ & $E=9$ \\ \midrule \midrule
FedGaLA & 67.8 & 65.2 & 64.8 & 64.7 \\ \bottomrule
\end{tabular}
\end{adjustbox}
\caption{Impact of communication frequency ($E$) on model accuracy over 900 total local epochs.}
\label{tab:comm_vs_perf}
\end{table}

\begin{table}[t]

\begin{center}
\resizebox{1\columnwidth}{!}{
\begin{tabular}{lccccc}
\toprule
Models & P & A & C & S & Ave. \\
\midrule \midrule
RF: 0.001 & 64.3(1.9) & 41.8(1.2) & 58.1(2.3) & 68.3(1.1) & 58.1 \\
RF: 0.01  & 63.8(1.4) & 42.0(1.7) & 58.1(0.4) & 68.3(1.6) & 58.6 \\
RF: 0.1   & 64.1(1.7) & 40.9(2.1) & 58.5(1.4) & 70.4(0.4) & 58.5 \\
FedGaLA (ours)          & \textbf{64.7(1.9)} & \textbf{44.2(1.2)} & \textbf{60.5(2.2)} & \textbf{70.5(1.3)} & \textbf{60.0} \\
\bottomrule
\end{tabular}
}
\end{center}
\caption{Results on PACS dataset for different reweight-factors and our FedGaLA method. `RF' indicates the re-weight factor.}
\label{Tab:reweight_results}
\end{table}

\begin{table*}[t]
\centering

\resizebox{0.9\linewidth}{!}{
\begin{tabular}{l|ccccccccc}
\toprule
\multirow{2}{*}{Model}& \multicolumn{5}{c|}{\textit{PACS}} & \multicolumn{4}{c}{\textit{DomainNet}}\\  
 & P & A & C & S & \multicolumn{1}{c|}{Ave.} & C & I & Q & Ave. \\ \midrule \midrule

FedGaLA+L2 ($\lambda=0.001$) & 64.8(0.5) & 42.8(1.1) & 58.7(1.6) & 67.3(0.6) & \multicolumn{1}{c|}{58.4} & 46.3(0.8) & \textbf{16.4(0.9)} & \textbf{62.8(0.2)} & \textbf{41.6} \\
FedGaLa+L2 ($\lambda=0.01$)  &  66.5(0.9)& 43.4(0.9) & 56.9(2.7) & 65.2(0.6) &  \multicolumn{1}{c|}{57.9} & 38.9(1.2) & 14.6(0.7) & 52.5(0.8) & 35.3 \\
FedGaLA+FedProx ($\mu=0.001$) & 62.1(0.2) & 41.4(0.7) & 58.9(3.1) & 68.8(0.1) & \multicolumn{1}{c|}{57.8} & 44.6(0.2) & 13.4(1.0) & 60.9(1.1) &39.6 \\ 
FedGaLA+FedProx ($\mu=0.01$) & 63.1(0.2) & 41.6(1.2) & 58.5(2.0) & 68.9(1.9) & \multicolumn{1}{c|}{58.0} & 44.7(1.2) & 13.2(0.4) & 61.7(0.4) & 39.9 \\ 
FedGaLA (ours) & \textbf{64.7(1.9)} & \textbf{44.2(1.2)} & \textbf{60.5(2.2)} & \textbf{70.5(1.3)} & \multicolumn{1}{c|}{\textbf{60.0}} & \textbf{47.6(0.9)} & 14.2(0.5) & 61.4(0.3) & 41.1 \\ \bottomrule
\end{tabular}%
}
\vspace{1mm}

 \centering
 \resizebox{0.95\linewidth}{!}{
\begin{tabular}{l|cccccccccc}
\toprule
\multirow{2}{*}{Model} & \multicolumn{5}{c|}{\textit{Office-Home}} & \multicolumn{5}{c}{\textit{TerraInc}} \\ 
 & \multicolumn{1}{c}{A} & C & P & R & \multicolumn{1}{c|}{Ave.} & L38 & L43 & L46 & L100 & Ave. \\ \midrule \midrule
 
FedGaLA+L2 ($\lambda=0.001$) &  \textbf{11.8(0.4)} & 23.9(0.5) & \textbf{37.1(0.4)} & \textbf{21.9(0.7)} & \multicolumn{1}{c|}{\textbf{23.6}} & 61.6(0.9) & 45.8(0.6) & 46.6(0.7)&70.8(0.7) & 56.2\\
FedGaLa+L2 ($\lambda=0.01$)  & \textbf{11.8(0.2)} & 21.2(0.1) & 34.5(0.3) & 22.1(0.2)& \multicolumn{1}{c|}{22.4} & 61.5(0.7) & 38.1(2.9) & 44.9(0.1) & 57.8(3.3) & 50.6\\
FedGaLA+FedProx ($\mu=0.001$) & 10.2(0.4) & 24.2(0.4) & 36.4(1.1) & 19.7(0.32) & \multicolumn{1}{c|}{22.6} & \textbf{63.8(0.7)} & \textbf{49.3(1.5)} & \textbf{47.2(0.1)} & 71.5(0.9) &\textbf{57.9} \\
FedGaLA+FedProx ($\mu=0.01$) & 9.2(0.3) & 23.6(0.7) & \textbf{37.1(0.6)} & 20.4(0.4) & \multicolumn{1}{c|}{22.6} & 63.2(1.1) & 47.8(2.1) & 44.8(0.4) & 71.4(0.8) & 56.0 \\
FedGaLA (ours) & 8.9(0.4) & \textbf{25.3(0.6)} & 36.6(0.3) & 21.2(0.5) & \multicolumn{1}{c|}{23.0} & 63.6(0.1) & 47.6(1.4) & 43.9(2.9) & \textbf{71.7(0.9)} & 56.7 \\ \hline
\end{tabular}%
}
\caption{ The effect of regularizers on the performance of FedGaLA across PACS, DomainNet, Office-Home, and TerraInc. }
\label{tab:fedprox_1}
\end{table*}

\subsection{Hyperparameter sensitivity and stability}

\noindent {\textbf{Batch size.}
To identify the optimal training hyperparameters, we conduct a grid search on various batch sizes, testing values of 32, 64, 128, and 256, as illustrated in Figure \ref{fig:batchh}. The results indicate that FedGaLA achieves its best performance with a batch size of 128, which balances efficient gradient updates and stability during training.

\noindent \textbf{Local threshold vs. the number of local epochs.} In Figure \ref{fig:TouE} we study the impact of the threshold for the similarity between gradients and the reference ($\tau$) and the number of local epochs $E$ on performance. In this experiment, we use three different values for $\tau$: $-$0.1, 0, and 0.1, and three different values for $E$: 5, 7 or 9 (local epochs per communication round).  
We observe that our method produces the best results when $\tau = 0$, i.e., when we keep all gradients with positive cosine similarity with the reference. Expectedly, even discarding gradients with small amounts of alignment ($\tau = 0.1$) degrades the results, while keeping gradients that are not aligned with the reference ($\tau=-0.1$) also hurts performance.   
Moreover, we see that setting $E = 7$ yields the best performance as increasing the number of epochs beyond 7 does not have a positive impact, and only increases computational time.

\noindent {\textbf{Global gradient alignment iterations.} Figure \ref{fig:iteration} shows the impact of different gradient alignment iterations (line 7 in Algorithm \ref{alg:UFDG}) on performance. FedGaLA produces the best results when the iteration number is 3. When the number of iterations increases, the model places too much emphasis on clients aligned with the global average and decreases the impact of unaligned clients harshly, leading to a decline in the overall performance of the model. On the other hand, with small iteration values, global gradient alignment is not performed effectively, and thus performance declines. 

\noindent {\textbf{Training stability.} 
 Figure \ref{fig:stability} demonstrates that FedGaLA maintains robust training stability, rapidly improving accuracy in the initial iterations and stabilizing after approximately 40 iterations. Despite discarding dissimilar gradients, the method avoids divergence and ensures consistent optimization, effectively learning domain-invariant features while maintaining steady performance throughout training.

\subsection{Data and communication efficiency}
\noindent \textbf{Ratio of labeled data.}
Figure \ref{fig:labeled} presents the results when evaluated with different ratios of labeled data, ranging from 10\% to 60\%. As can be seen, the accuracy increases significantly from 64\% to 72\% when the label ratio rises from 10\% to 30\%, followed by a steady climb to 74\% as the label ratios increase to 60\%. Overall, when more labeled data are used to train the linear classifier, the final performance expectedly improves.

\noindent \textbf{Communication rounds.} We investigate the effect of the communication rounds $T$ on the model's performance. Following \cite{zhuang2021collaborative,zhuang2022divergence} we set $E=1$ and vary the number of communication rounds $T$ from 100 to 900. We observe from Figure \ref{fig:T} that performance improves significantly from 1 to 200 communication rounds, with this trend slowing down from 200 to 900. 


\noindent\textbf{FedGaLA and communication cost.}
We conduct an experiment to analyze the trade-off between communication frequency and model performance. In this setup, local models are trained for a total of 900 epochs, with aggregation occurring after every $E$ local epochs. As $E$ increases, communication frequency decreases, thereby reducing the overhead. The performance of the model is then assessed for different values of $E$, as demonstrated in Table~\ref{tab:comm_vs_perf}.
The results indicate that increasing the communication frequency (lower $E$) improves accuracy and the model maintains consistent performance even with reduced communication, particularly for $E > 1$. This indicates the ability of the model to adapt to less frequent aggregation intervals.

\subsection{Model variations}

\noindent \textbf{Gradient discarding vs. gradient re-weighting.}
Gradient re-weighting is an alternative approach to discarding unaligned gradients. To further highlight the effectiveness of gradient discarding in the local alignment module of FedGaLA, we conducted an additional experiment replacing gradient discarding with gradient re-weighting in FedGaLA. Here, rather than discarding the unaligned gradients, they are multiplied by a re-weight factor reducing their impact. The results are presented in  Table~\ref{Tab:reweight_results}, demonstrating that while re-weighting is effective, our original method of discarding unaligned gradients yields the best overall performance.

\noindent \textbf{Effects of regularizers on FedGaLA.} 
Prior works have demonstrated that adding regularizers can indeed improve generalization across domains or non-IID data \cite{li2020federated, nguyen2022fedsr}. 
To this end, we test the impact of regularizers on FedGaLA by applying two types of regularizers based on L2 norm \cite{li2020federated} and FedProx \cite{nguyen2022fedsr}. Please refer to Appendix \ref{app:regs} for more details regarding these two techniques. The results in Table \ref{tab:fedprox_1} demonstrate that FedGaLA is highly compatible with regularizers and that the addition of such approaches can further boost the performance of our method.

\section{Conclusion}
\label{sec:conclusion}
In this work, we first introduced a new problem category, federated unsupervised domain generalization, and subsequently proposed FedGaLA to tackle it. Building the relationship between gradient alignment and domain shift, FedGaLA comprises two alignment strategies at the global and local levels designed to address the problem of domain generalization in an unsupervised and federated setup. We assessed the performance of FedGaLA through extensive experiments where our approach outperformed various baseline models. Detailed ablation studies and sensitivity analyses were also conducted to provide more insights into different aspects of our solution.

\section*{Acknowledgments}
We would like to thank Geotab Inc., the City of Kingston, and NSERC for their support of this work.

\bibliography{aaai25}

\newpage

\appendix


\section{Proofs}
\label{app:proof}
\subsection{Proof of Lemma 1} 
\label{proof_lemma_0}
\begin{proof}
By definition, the Mutual Information between two random variables, $\textbf{x}_i$ and $\textbf{x}_j$ is
\begin{align}\label{eq:def}
I(\textbf{x}_i; \textbf{x}_j) = \int \int p(\textbf{x}_i, \textbf{x}_j) \log\left(\frac{p(\textbf{x}_i, \textbf{x}_j)}{p(\textbf{x}_i)p(\textbf{x}_j)}\right) d\textbf{x}_i d\textbf{x}_j.
\end{align}
Since the domains from which the corresponding random variables are drawn are sets of independent features:
\begin{equation}\label{eq:p_x}
p(\textbf{x}_i) = \prod_{f=1}^Fp(x_i^f), \hspace{0.5cm} p(\textbf{x}_j) = \prod_{f=1}^Fp(x_j^f).
\end{equation}
Moreover, since each pair of corresponding features ($x_i^f,x_j^f$) across domains forms a bivariate Normal distribution, we can derive:
\begin{equation}\label{eq:p_xy}
p(\textbf{x}_i,\textbf{x}_j) = \prod_{f=1}^Fp(x_i^f,x_j^f).
\end{equation}
By substituting Eqs. (\ref{eq:p_x}) and (\ref{eq:p_xy}) into Eq. (\ref{eq:def}), using the multiplicative property of logarithms and the definition of Mutual Information, we obtain:
\begin{multline}\label{I_featuresss}   
I(\textbf{x}_i; \textbf{x}_j) =\\ \sum_{f=1}^F\int \int p(x_i^f, x_j^f) \log\left(\frac{p(x_i^f, x_j^f)}{p(x_i^f)p(x_j^f)}\right) dx_i^f dx_j^f  
= \\ \sum_{f=1}^FI(x_i^f; x_j^f).
\end{multline}
For bivariate Normal distributions $x_i$ and $x_j$, Mutual Information can be measured by $I(x_i; x_j) = -\frac{1}{2} \log(1 - \rho^2)$, where $\rho = \frac{\sigma_{x_i,x_j}}{\sigma_{x_i} \sigma_{x_j}}$ is the correlation coefficient \cite{goldfeld2021sliced}.
Accordingly, Eq. (\ref{I_featuresss}) yields:
\begin{equation} \label{eq8}
I(\textbf{x}_i; \textbf{x}_j) = -\frac{1}{2}\sum_{f=1}^F\log(1-(\frac{\sigma_{x_i^f,x_j^f}}{\sigma_{x_i^f}\sigma_{x_j^f}})^2).
\end{equation}
According to Assumption 1, we have $\sigma_{x_i^f}=\sigma_{x_j^f}=1$. Hence, Eq.~(\ref{eq8}) yields:
\begin{equation} \label{eq12}
I(\textbf{x}_i; \textbf{x}_j) = -\frac{1}{2}\sum_{f=1}^F\log(1-\sigma_{x_i^f,x_j^f}^2),
\end{equation}
which completes the proof.
\end{proof}

\subsection{Proof of Lemma 2} 
\label{Ap:lem2}

\begin{proof}
The covariance of $\textbf{g}_i$ and $\textbf{g}_j$, by definition, is:

\begin{equation}
     \text{Cov}(\textbf{g}_i,\textbf{g}_j) = \mathbb{E}[(\textbf{g}_i-\mathbb{E}[\textbf{g}_i])(\textbf{g}_j-\mathbb{E}[\textbf{g}_j])^T].
\end{equation}
First, we calculate $\mathbb{E}[\textbf{g}_i]$. Given $\textbf{g}$ is differentiable, as per the Assumption 1, we can estimate it using the first-degree Taylor expansion theorem around $\mu$:
\begin{equation}\label{eq2}
     g(x) \approx g(\mu) + J_g(\mu) . (x - \mu),
\end{equation} 
where $J_g(.)$ is the Jacobian matrix $g$. We can use this equation to estimate $\mathbb{E}[\textbf{g}]$ as follows:
\begin{equation}
     \mathbb{E}[\textbf{g}] = \mathbb{E}\left[g(\mu) +  J_g(\mu). (x - \mu))\right].
\end{equation} 
Note that $g(\mu)$ is constant, therefore $\mathbb{E}[g(\mu)] = g(\mu)$. Moreover, since we assume that the distribution of the features is normal, we can conclude: 
\begin{equation}
     \mathbb{E}\left[ J_g(\mu). (x - \mu))\right] = \textbf{0}_F.
\end{equation} 
Since $\mathbb{E}[x - \mu]=\textbf{0}_F$ for $x \sim \mathcal{N}$, $\mathbb{E}[g] = g(\mu)$. By replacing $\textbf{g}_i$ and $\textbf{g}_j$ with their Taylor expansion we derive:
\begin{multline}
     \text{Cov}(\textbf{g}_i,\textbf{g}_j) = \\ \mathbb{E}\left[ \left(J_{g_i}(\mu_i). (x - \mu_i)\right) 
                             \left. \right. . \left(J_{g_j}(\mu_j). (x - \mu_j))\right)^T \right].
\end{multline}
Hence:
\begin{multline}
     \text{Cov}(\textbf{g}_i,\textbf{g}_j) = \\  J_{g_i}(\mu_i).\mathbb{E}\left[ (x - \mu_i) 
                              .(x - \mu_j)^T \right] . J_{g_j}(\mu_j)^T .
\end{multline}
Given the covariance matrix $\Sigma = \mathbb{E}[(x-\mu_i)(x-\mu_j)] $, we derive:
\begin{align}
     \text{Cov}(\textbf{g}_i,\textbf{g}_j) = &  J_{g_i}(\mu_i).\Sigma . J_{g_j}(\mu_j)^T .
\end{align}
As the features are assumed to be independent, $\Sigma$ is a diagonal matrix, the general term of each entry of $\text{Cov}(\textbf{g}_i,\textbf{g}_j)$ can be derived as:
\begin{multline}
     \text{Cov}(\textbf{g}_i,\textbf{g}_j)_{mn} =  \\ \sum_{f=1}^F\sigma_{x^f_i,x^ f_j}\left(\frac{\partial g_{i(m)}}{\partial x_i^f}\bigg|_{\mu_i^f}\right)
                            \left(\frac{\partial g_{j(n)}}{\partial x_j^f}\bigg|_{\mu_j^f}\right),
\end{multline}
where $g_{i(m)}$ and $g_{j(n)}$ represented the $m^{th}$ and the $n^{th}$ dimension of $g_i$ and $g_j$, respectively. 
\end{proof}

\subsection{Proof of Claim 1} 
\label{proof_claim_2}
\begin{proof}
Following Assumption 1, the model is defined as $\hat{y}=\sigma(Wx_1-Wx_2)$ and the loss is formulated as:
\begin{equation} 
l(y, \hat{y}) = -y \log(\hat{y}) - (1 - y) \log(1 - \hat{y}),
\end{equation}
where $y=1$ for positive pairs and $y=0$ for the negative pairs. Consequently, the gradient of the loss w.r.t. $W$ is
\begin{multline}\label{main_deriv}
g = \\ \left\{
\begin{array}{ll}
(\sigma(Wx_1-Wx_2)-1)(x_1-x_2) & \text{if } y=1 \\ & \text{ (Positive pairs)} \\
\sigma(Wx_1-Wx_2)(x_1-x_2) & \text{if } y=0 \\ & \text{ (Negative pairs)}\\
\end{array}
\right.
\end{multline}
We first focus on the positive pairs. In the contrastive loss, an augmentation of the same sample is usually used as the positive sample. Following Assumption 1, the augmented sample is formulated as ${x}_2=Ax_1+B$. Hence, the gradient of the positive pair is derived as:
\begin{equation}
g^+ = 
(\sigma(Wx_1-WAx_1-WB)-1)(x_1-Ax_1-B).
\end{equation}
Subsequently, the derivative of $g^+$ with respect to $x_1$ at $x_1 = \mu_1$ is expressed as:
\begin{multline}\label{I_features}   
\frac{\partial {g^+}}{\partial x_1}\bigg|_{x_1=\mu_1} = \\ \sigma(W\mu_1-WA\mu-WB)(1-\sigma(W\mu_1-WA\mu_1-WB)) \\ (\mu_1-A\mu-B) \nonumber  
  + \sigma(W\mu_1-WA\mu_1-WB)(1-A). 
\end{multline}
Recall Assumption 1 stating that the data is normalized, hence, $\mu_1=0$. Therefore: 
\begin{multline}
\frac{\partial {g^+}}{\partial x_1}\bigg|_{x_1=\mu_1}  = \\ \sigma(-WB)(1-\sigma(-WB)) (-B) 
 + \sigma(-WB)(1-A). 
\end{multline}
As a result, we can conclude that for all positive pairs, $\frac{\partial {g^+}}{\partial x_i}\bigg|_{x_i=\mu_i}$ only depends on the augmentation and the network weights. According to Assumption 1, the weights of clients are the same and equal to the global model at each communication round. Moreover, since the same random augmentations are broadcasted to all the clients,  $A$ and $B$ will be the same at each step across all the local models. Hence, for all $x_i$, $\text{sign}(\frac{\partial {g}_i}{\partial x^f_i}\big|_{x^f_i=\mu^f_i}) = \text{sign}(\frac{\partial {g}_j}{\partial x^f_j}\big|_{x^f_j=\mu^f_j}).$

Considering the negative pairs, from Eq. (\ref{main_deriv}), we have:
\begin{equation}
g^- = 
\sigma(Wx_1-Wx_2)(x_1-x_2). 
\end{equation}
The derivative of $g^-$ w.r.t. $x_1$ at $x_1 = \mu_1$ is derived as:
\begin{multline}\label{I_fea}   
\frac{\partial {g^-}}{\partial x_1}\bigg|_{x_1=\mu_1}  =  \sigma(W\mu_1-Wx_2)(1-\sigma(W\mu_1-Wx_2)) \\ 
(\mu_1-Ax_2) + \sigma(W\mu_1-Wx_2)(1-x_2).
\end{multline}
Based on the stated assumption $\mu_1=0$, therefore:
\begin{multline} \label{g_neg}
\frac{\partial {g^-}}{\partial x_1}\bigg|_{x_1=\mu_1} = \sigma(-Wx_2)(1-\sigma(-Wx_2)) \\ (-Ax_2)  + \sigma(-Wx_2)(1-x_2). 
\end{multline}
The $\sigma$ function is estimated using Taylor expansion around zero (since the data is normalized to $\mu=0$), as:
\begin{equation}\label{sigma_taylor}
\sigma(x) \simeq \frac{1}{2} + \sum_{n=0}^\infty \frac{(-1)^n (2^{2n+1} - 1) B_{2n+2}}{(2n+2)!} x^{2n+1},
\end{equation}
where $B_n$ are Bernoulli numbers. 
By replacing Eq. (\ref{sigma_taylor}) in Eq. (\ref{g_neg}):
\begin{small}
\begin{multline}
\frac{\partial g^-}{\partial x_1}\bigg|_{x_1=\mu_1} \simeq \\ \left(\frac{1}{2} + \sum_{n=0}^\infty \frac{(-1)^{n+1} (2^{2n+1} - 1) B_{2n+2}}{(2n+2)!} (Wx_2)^{2n+1}\right) \quad \times \\ \left(\frac{1}{2} - \sum_{n=0}^\infty \frac{(-1)^{n+1} (2^{2n+1} - 1) B_{2n+2}}{(2n+2)!} (Wx_2)^{2n+1}\right)(-Ax_2) \\  
\quad + \left(\frac{1}{2} + \sum_{n=0}^\infty \frac{(-1)^{n+1} (2^{2n+1} - 1) B_{2n+2}}{(2n+2)!} (Wx_2)^{2n+1}\right)(1-x_2). \\ 
\end{multline}
\end{small}
Since the coefficients are only consist of $n$, $B_n$ and $W$, we can simplify as:
\begin{equation}\label{sigma_taylorrr}
\frac{\partial {g^-}}{\partial x_1}\bigg|_{x_1=\mu_1} \approx \sum_{n=0}^\infty \zeta(W,n)x_2^n
\end{equation}

Since the gradient is a linear operation, the gradient of the sum equals the sum of gradients. According to Assumption \ref{assumption2}, we update the model with the loss of the entire dataset as the negative samples. Hence:
\begin{equation}\label{sigma_taylor_sum}
\frac{\partial {g^-}}{\partial x_1}\bigg|_{x_1=\mu_1} = \sum_{i=2}^N \sum_{n=0}^\infty \zeta(W,n)x_i^n.
\end{equation}
By substituting the sums, we derive:
\begin{equation}\label{sigma_taylor_sum_sub}
\frac{\partial {g^-}}{\partial x_1}\bigg|_{x_1=\mu_1} =  \sum_{n=0}^\infty \sum_{i=2}^N\zeta(W,n)x_i^n.
\end{equation}
Since the distribution of data is assumed to be normal, these sums can be estimated using the expected value of $\mathbb{E}(\sum_{i=1}^N x_i^n) = N\mathbb{E}(x^n)$. Since $x$ belongs to a normal distribution with $\mu=0$ and $\sigma=1$, the sum can be estimated as:
\begin{equation}
\sum_{i=1}^N x^n \simeq \left\{
\begin{array}{cll}
(n-1)!! &  n = 2k& \text{ (Even powers)} \\
0 &  n=2k+1 &\text{ (Odd powers)}
\end{array}
\right.
\end{equation}
By applying this estimate to Eq. ( \ref{sigma_taylor_sum}), we derive:
\begin{equation}\label{sigma_taylor_sum2}
\frac{\partial {g^-}}{\partial x_1}\bigg|_{x_1=\mu_1} =  \sum_{k=0}^\infty \sum_{i=2}^N\zeta(W,n)(n-1)!!
\end{equation}
Consequently, we can conclude that for all negative pairs, 
the gradient derivative 
$\frac{\partial {g^-}}{\partial x_i}\bigg|_{x_i=\mu_i}$ only depends on the network weights. Hence, due to the fact that both matrices are symmetric,  $\text{sign}(\frac{\partial {g}_i}{\partial x^f_i}\big|_{x^f_i=\mu^f_i}) = \text{sign}(\frac{\partial {g}_j}{\partial x^f_j}\big|_{x^f_j=\mu^f_j})$.
    
\end{proof}

\subsection{Federated supervised domain generalization}\label{proof:supervised}

\renewcommand{\theremark}{A\arabic{remark}} 
\setcounter{remark}{0} 

\begin{remark}\label{remark}
The insights from Theorem 1 also apply to the federated \textbf{supervised} learning framework, demonstrating that an increase in domain shift leads to a monotonic decrease in the covariance of the gradients of local models within the context of federated supervised domain generalization.
\end{remark}

We prove this remark under the following assumptions, which differ slightly from those previously presented in Assumption 1:

\renewcommand{\theassumption}{A\arabic{assumption}} 
\setcounter{assumption}{0} 

\renewcommand{\theclaim}{A\arabic{claim}} 
\setcounter{claim}{0} 

\begin{assumption}\label{assumption2}
Based on Assumption 1, we extend our framework to a supervised learning context. The assumptions regarding the differentiability of the gradients and data distribution remain the same except the clients now do have access to the labels. Regarding the model,  we assume each client is a logistic regression classifier trained in a supervised federated setup using cross-entropy loss.
\end{assumption}

\begin{proof}
Since all the assumptions regarding the data distribution are the same, Eq. (\ref{eq:def}) to Eq. (4) still hold. However, since the training paradigm has changed, we introduce the following claim, which extends Claim 1 under the conditions of Assumption \ref{assumption2}.

\begin{claim} \label{claim1}
For all clients with a logistic regression classifier described in Assumption \ref{assumption2}, for any $i$ and $j$, the sign of $(\frac{\partial \textbf{g}_i}{\partial x^f_i}\bigg|_{x^f_i=\mu^f_i})(\frac{\partial \textbf{g}_j}{\partial x^f_j}\bigg|_{x^f_j=\mu^f_j})$ is always positive. For the proof see Appendix \ref{proof_claim_1}. 
\end{claim}

From Eq. (4) and Claim \ref{claim1} it can be concluded $\text{Cov}(\textbf{g}_i,\textbf{g}_j)$ and $I(\textbf{x}_i; \textbf{x}_j)$ are positively and monotonically correlated, which completes the proof.

\end{proof}

\subsection{Proof of Claim \ref{claim1}}\label{proof_claim_1}
\begin{proof}

Following the assumptions, let's describe the model's loss by:
\begin{equation}
l(y, \hat{y}) = -y \log(\hat{y}) - (1 - y) \log(1 - \hat{y}),
\end{equation}
where $y$ represents the label and $\hat{y}$ signifies the model output defined as $\sigma(Wx+b) =\sigma(x)=1/(1+e^{-Wx+b})$. The gradient of the loss with respect to $W$ is therefore given by: 
\begin{equation}
g = \frac{\partial L}{\partial w} = (\hat{y} - y)x,
\end{equation}
Subsequently, the derivative of $g$ with respect to $x$ at $x=\mu$ is expressed as:
\begin{equation} 
\frac{\partial {g}}{\partial x}\bigg|_{x=\mu} = \sigma(W\mu+b) +  \mu\frac{\partial \sigma(Wx+b)}{\partial x}\bigg|_{x=\mu}-y.
\end{equation}
Recalling the proof of theorem 1 and noting that data is standardized, we find $\mu_i^f=\mu_j^f = 0$. This leads to
\begin{equation} 
\frac{\partial {g}}{\partial x}\big|_{x=\mu} = \sigma(b) -y.
\end{equation}
Given that $\sigma(x) = 1/(1+e^{-x})$, its outcome always falls within the range $(0,1)$. Considering $y$ as a data label that can be either 1 or 0, it follows that when $y=1$, $\sigma(b) -y$ is invariably negative, and when $y=0$, $\sigma(b) -y$ is invariably positive. Consequently, the sign of $\frac{\partial \mathbf{g}}{\partial x}\big|_{x=\mu}$ is determined solely by the value of $y$, ensuring $\text{sign}(\sigma(b_i) -y) = \text{sign}(\sigma(b_j) -y)$. Thus,
\begin{equation} 
\text{sign}(\frac{\partial {g}_i}{\partial x^f_i}\big|_{x^f_i=\mu^f_i}) = \text{sign}(\frac{\partial {g}_j}{\partial x^f_j}\big|_{x^f_j=\mu^f_j}).
\end{equation}
    
\end{proof}

\subsection{Proof of Corollary 1}
\label{Ap:rem1}
\begin{proof}
    
The variance of $\textbf{g}_i$ and $\textbf{g}_j$ is:
\begin{equation}
    \text{Var}(\textbf{g}_i-\textbf{g}_j)=\text{Var}(\textbf{g}_i)+\text{Var}(\textbf{g}_j)-2\text{Cov}(\textbf{g}_i,\textbf{g}_j).
\end{equation}
For deriving $\text{Var}(\textbf{g}_i)$ and $\text{Var}(\textbf{g}_j)$, we first introduce Lemma \ref{lemma_2}.

\renewcommand{\thelemma}{A\arabic{lemma}} 
\setcounter{lemma}{0} 

\begin{lemma}\label{lemma_2}
Given the assumptions stated, the variance of a differentiable function $\textbf{g}$ can be estimated as:
\begin{equation}
    \text{Var}(\textbf{g}) = \sum_{f=1}^F(\sigma^f)^2(\frac{\partial {g}}{\partial x^f}\bigg|_{x^f=\mu^f})^2,
\end{equation}
where $x^f$ is the $f^{th}$ feature of $\textbf{x}$, $\sigma^f$ is its standard deviation, and $\mu^f$ is its mean. The full proof of this lemma is presented in Appendix \ref{Ap:lem3}.      
\end{lemma} 

By derive $\text{Var}(\textbf{g}_i)$ and $\text{Var}(\textbf{g}_j)$ using Lemma 2 and $\text{Cov}(\textbf{g}_i,\textbf{g}_j)$ from Lemma 1, we derive:
\begin{multline}\label{equ:6}
\text{Var}(\textbf{g}_i-\textbf{g}_j) = \\ \sum_{f=1}^F(\sigma^f_i)^2(\frac{\partial {g}_i}{\partial x^f_i}\bigg|_{x^f_i=\mu^f_i})^2 
 + (\sigma^f_j)^2(\frac{\partial {g}_j}{\partial x^f_j}\bigg|_{x^f_j=\mu^f_j})^2 
\\ - 2\sigma_{x_i^f,x_j^f}(\frac{\partial {g}_i}{\partial x^f_i}\bigg|_{x^f_i=\mu^f_i})(\frac{\partial {g}_j}{\partial x^f_j}\bigg|_{x^f_j=\mu^f_j}).
\end{multline}
If we assume two domains that have the same distribution but different covariances with $\textbf{g}_i$ denoted by $\textbf{g}_{j1}$ and $\textbf{g}_{j2}$, then:
\begin{equation}\label{equ:7}
    \text{Var}(\textbf{g}_i-\textbf{g}_{j1})-\text{Var}(\textbf{g}_i-\textbf{g}_{j2}) =\sum_{f=1}^F \alpha_f (\sigma^{f}_{ij1}-\sigma^{f}_{ij2}),
\end{equation}
where $\alpha_f = 2(\frac{\partial {g}_i}{\partial x^f_i}\bigg|_{x^f_i=\mu^f_i}) ( \frac{\partial {g}_{j}}{\partial x^f_j}\bigg|_{x^f_j=\mu^f_j})$ which is always positive (According to Claims \ref{claim1} and 1).
\end{proof}
This conclusion highlights our finding in Theorem 1 regarding the relationship between the distribution of different domains and their corresponding gradients. 
\subsection{Proof of Lemma \ref{lemma_2}}
\label{Ap:lem3}

\begin{proof}
The variance of $\textbf{g}$, by definition, is:
\begin{equation}\label{eq1}
     \text{Var}(\textbf{g}) = \mathbb{E}{(\textbf{g}-\mathbb{E}{\textbf{g}})^2}.
\end{equation} 
From the proof of Lemma 2, we know that $\mathbb{E}{g} = g(\mu)$. By expanding $g$ around $\mu$ we derive:
\begin{equation}
    \text{Var}(\textbf{g}) = \mathbb{E}\left[\left(\sum_{f=1}^F (x^f - \mu^f)\left(\frac{\partial g}{\partial x^f}\bigg|_{x^f=\mu^f}\right)\right)^2\right].
\end{equation} 
The squared term can be expanded as:
\begin{multline}
    \text{Var}(\textbf{g})=  \\ \mathbb{E}[\sum_{f=1}^F \sum_{e=1}^F (x^f - \mu^f)(x^e - \mu^e)\left(\frac{\partial g}{\partial x^f}\bigg|_{x^f=\mu^f}\right) \\
     \left(\frac{\partial g}{\partial x^e}\bigg|_{x^e=\mu^e}\right)  ].
\end{multline} 
As the features are assumed to be independent, when $f \neq e$, $\mathbb{E}{(x^f - \mu^f)(x^e - \mu^e)}$ represents $\text{Cov}(x^f, x^e)$ and equals zero. Therefore:
\begin{multline}
    \text{Var}(\textbf{g}) = \\ \mathbb{E}\left[\sum_{f=1}^F (x^f - \mu^f)^2\left(\frac{\partial g}{\partial x^f}\bigg|_{x^f=\mu^f}\right)^2\right].
\end{multline}
Since $\mathbb{E}{(x^f - \mu^f)^2} = (\sigma^f)^2$, we derive:
\begin{equation}
\text{Var}(\textbf{g}) = \sum_{f=1}^F (\sigma^f)^2\left(\frac{\partial g}{\partial x^f}\bigg|_{x^f=\mu^f}\right)^2,
\end{equation}
which completes the proof for the variance of $\textbf{g}$.
\end{proof}

\subsection{Proof of Proposition 1}\label{prop_proof}
\begin{assumption}
    Let $\textbf{\emph{g}}_{i}$ and $\textbf{\emph{g}}_{j}$ be two sets of gradients. $\textbf{\emph{g}}_{est}$ is defined as the average of both sets. We assume $cos(\textbf{\emph{g}}_{j,k},\textbf{\emph{g}}_{est})>0$ for all $k\neq K$, $cos(\textbf{\emph{g}}_{i,k},\textbf{\emph{g}}_{est})>0$ for all $k$, and $cos(\textbf{\emph{g}}_{j,K},\textbf{\emph{g}}_{est})<0$. Without the loss of generality, we assume the last vector in the set is the unaligned vector.
    \end{assumption}
\begin{proof}
We first define $\hat{\textbf{g}_{j}}$ as:
\begin{equation}
    \textbf{g}_{j} = \hat{\textbf{g}_{j}} \cup \{\textbf{g}_{j,K}\}.
\end{equation}
Hence, the mean of $\textbf{g}_{j}$ can be derived as:
\begin{multline}
    \mu_{\textbf{g}_{j}} = \\ \frac{1}{K} ((K-1) \mu_{\hat{\textbf{g}_{j}}} + \textbf{g}_{j,K}), \quad \mu_{\hat{\textbf{g}_{j}}} = \frac{1}{K-1}\sum_{k=1}^{K-1}\textbf{g}_{j,K}.
\end{multline}
Now we shift to the covariance of the two sets:
\begin{equation}\label{cov_def}
    \text{Cov}(\textbf{g}_{i}, \textbf{g}_{j}) = \frac{1}{K-1} (\textbf{g}_{i} - \mu_{\textbf{g}_{i}})(\textbf{g}_{j} - \mu_{\textbf{g}_{i}})^T.
\end{equation}
Substituting $\text{Cov}(\textbf{g}_{i}, \hat{\textbf{g}_{j}})$ in Eq. \ref{cov_def}, we derive:
\begin{multline}
    \text{Cov}(\textbf{g}_{i}, \textbf{g}_{j}) = \\ \frac{K-2}{K-1} \text{Cov}(\textbf{g}_{i}, \hat{\textbf{g}_{j}}) + \frac{1}{K-1}(\textbf{g}_{i} - \mu_{\textbf{g}_{i}})(\textbf{g}_{j,k} - \mu_{\textbf{g}_{i}})^T.
\end{multline}
By substituting $\mu_{\textbf{g}_{j}}$ in this equation we have:
\begin{multline}
    \text{Cov}(\textbf{g}_{i}, \textbf{g}_{j}) = \\ \frac{K-2}{K-1} \text{Cov}(\textbf{g}_{i}, \hat{\textbf{g}_{j}}) + \frac{1}{K-1}(\textbf{g}_{i} - \mu_{\textbf{g}_{i}}) \\ (\textbf{g}_{j,k} - \frac{1}{K} ((K-1) \mu_{\hat{\textbf{g}_{j}}} + \textbf{g}_{j,K}))^T.
\end{multline}
Simplifying this equation we derive:
\begin{multline}
    \text{Cov}(\textbf{g}_{i}, \textbf{g}_{j}) = \\ \frac{K-2}{K-1} \text{Cov}(\textbf{g}_{i}, \hat{\textbf{g}_{j}}) + \frac{1}{K}(\textbf{g}_{i} - \mu_{\textbf{g}_{i}})  (\textbf{g}_{j,K} -  \mu_{\hat{\textbf{g}_{j}}})^T.
\end{multline}
The term $(\textbf{g}_{i} - \mu_{\textbf{g}_{i}})  (\textbf{g}_{j,K} -  \mu_{\hat{\textbf{g}_{j}}})^T$ is a matrix with mostly negative components because of the following reasons. $(\textbf{g}_{i} - \mu_{\textbf{g}_{i}})$ represents the deviations of $\textbf{g}_{i}$ vectors from their mean which points around $\textbf{g}_{est}$ as $cos(\textbf{g}_{i,k},\textbf{g}_{est})>0$ for all $k$. On the other hand, since $cos(\textbf{g}_{j,K},\textbf{g}_{est})<0$, $(\textbf{g}_{j,K} -  \mu_{\hat{\textbf{g}_{j}}})^T$ representing the deviation of $\textbf{g}_{j,K}$ from the mean vector is in the opposite direction of $\textbf{g}_{est}$. Hence, the dot product of two sets of vectors pointing in two opposite directions results in a negative covariance matrix which completes the proof.
\end{proof}

\section{Implementation details}
\label{app:imp}

In this section, we provide further implementation details including image augmentation, network architecture, evaluation approach, and regularizers.

\subsection{Image augmentations} We follow  \cite{chen2020simple} for augmentations. For all datasets, a random patch of the image is selected and resized to $32\times32$. Subsequently, we apply two random transformations, namely horizontal flip and color distortion.

\subsection{Network architecture}
\label{SecSec:arch}

We use ResNet18 \cite{he2016deep} as the encoder for all the experiments. We use the ResNet architecture presented in \cite{zhuang2022divergence}, which is slightly different from the original architecture: (\textit{i}) The first convolution layer employs a $3\times3$ kernel size, replacing the original $7\times7$; (\textit{ii}) An average pooling layer with a $4\times4$ kernel size is used before the final linear layer, substituting the adaptive average pooling layer; and (\textit{iii}) The last linear layer is replaced with a two-layer MLP, which shares the same network architecture as the predictor.

\subsection{Training and evaluation}
We implement all our models using PyTorch and provide an easy-to-use framework for federated domain generalization in our released repository. Below, we provide further details regarding the hyperparameters used in the training and evaluation processes.

\noindent \textbf{Training.} While training the clients, we use a batch size of 128. Each client is trained for 7 local epochs before being returned to the server for a communication round. By default, we train for 100 communication rounds. 
We use the Adam \cite{kingma2014adam} optimizer with a learning rate of $3\times10^{-3}$. The hyperparameters used for training FedSimCLR are the same as those used in FedGaLA. For other baselines (FedMoCo, FedSimSiam, FedBYOL, and FedEMA), we use the SGD optimizer with a momentum of 0.9 and a weight decay of $3\times10^{-4}$. The choice of learning rate for FedSimSiam, FedBYOL, and FedEMA is $0.03$ while for FedMoCo we use a learning rate of $0.025$. The parameters have been tuned to maximize performance.

\noindent \textbf{Evaluation.} Linear evaluation is used to assess the quality of learned representations. We train a fully connected layer on the top of the frozen encoder, which is trained for 100 epochs using the Adam optimizer with a learning rate $3\times10^{-3}$.

\subsection{Regularizers}
\label{app:regs}
Below we provide the details of the L2-norm and proximal term regularizers used in \cite{nguyen2022fedsr} and \cite{li2020federated}, respectively. 

\noindent \textbf{L2-norm.} Suppose $\Theta_i$ indicates the parameters of the $i^\text{th}$ client and $\mathcal{L}_{SSL,i}$ represents the self-supervised loss of the $i^\text{th}$ client. The L2-norm can be added to the loss of the $i^\text{th}$ client to obtain
\begin{align}
    \mathcal{L}_{\text{\textit{Total}},i}=\mathcal{L}_{SSL,i}+\lambda \|\Theta_i\|_2^2,
\end{align}
where $\lambda$ is the regularization coefficient and $\|\Theta_i\|_2^2$ is the
square of the L2-norm of the parameters, defined as:
\begin{align}
    \|\Theta_i\|_2^2 = \sum_{j} \theta_{i,j}^2,
\end{align}
where $\theta_{i,j}$ represents the  $j^\text{th}$ parameter of the  $i^\text{th}$ client.

\noindent \textbf{Proximal term.}
We also utilize the proximal term from \cite{li2020federated} to further penalize the deviation of local models from the global model. The formulation of the proximal term is:
\begin{align}
    \mathcal{L}_{\text{\textit{Total}},i} = \mathcal{L}_{SSL,i}+\mu \|\Theta_i - \Theta_g\|_2^2,
\end{align}
where $\mu$ is the regularizer coefficient, $\Theta_g$ and $\Theta_i$ are the parameters of the global model and $i^\text{th}$ client, respectively, and $\|\Theta_i - \Theta_g\|_2$ is the Euclidean norm of the difference of the weights of the local clients from the weights of the global model. This ensures that the parameters of the local model $\Theta_i$ do not diverge heavily from the previous global model.

\end{document}